%% file: paper_25.tex
\documentclass{article}
\usepackage{ijcai25}

\usepackage{times}
\usepackage{soul}
\usepackage{url}
\usepackage[hidelinks]{hyperref}
\usepackage[utf8]{inputenc}
\usepackage[small]{caption}
\usepackage{graphicx}
\usepackage{booktabs}
\usepackage[switch]{lineno}

\usepackage{amssymb}
\usepackage{multirow}
\usepackage{pifont}
\usepackage{subcaption}
\usepackage{amsmath}
\usepackage{amsthm}
\usepackage{booktabs}
\usepackage[linesnumbered,ruled,vlined]{algorithm2e}
\usepackage[switch]{lineno}
\usepackage{mdframed}
\usepackage[dvipsnames]{xcolor}
\usepackage{color, colortbl}

\definecolor{HighLight}{rgb}{0.96,0.92,0.96}

\title{\textsc{OT-detector}: Delving into Optimal Transport for Zero-shot Out-of-Distribution Detection}

\author{
Yu Liu$^{1}$
\and
Hao Tang$^{2}$
\and
Haiqi Zhang$^1$
\and
Jing Qin$^2$
\And
Zechao Li$^{1,*}$
\affiliations
$^1$School of Computer Science and Engineering, Nanjing University of Science and Technology\\
$^2$Centre for Smart Health, The Hong Kong Polytechnic University\\
\emails
\{yu.liu, zhanghq2017, zechao.li\}@njust.edu.cn,
\{howard-hao.tang, harry.qin\}@polyu.edu.hk
}

\begin{document}

\maketitle
\input{Sections/0-abstract.tex}
\input{Sections/1-introduction.tex}

\input{Sections/2-related.tex}

\input{Sections/3-method.tex}
\input{Sections/4-experiments.tex}
\input{Sections/5-conclusion.tex}


\bibliographystyle{named}
\bibliography{reference}
\clearpage
\input{Sections/Appendix}

\end{document}

%% file: Sections/0-abstract.tex
\begin{abstract}

    Out-of-distribution (OOD) detection is crucial for ensuring the reliability and safety of machine learning models in real-world applications. While zero-shot OOD detection, which requires no training on in-distribution (ID) data, has become feasible with the emergence of vision-language models like CLIP, existing methods primarily focus on semantic matching and fail to fully capture distributional discrepancies. To address these limitations, we propose \textsc{OT-detector}, a novel framework that employs Optimal Transport (OT) to quantify both semantic and distributional discrepancies between test samples and ID labels. Specifically, we introduce cross-modal transport mass and transport cost as semantic-wise and distribution-wise OOD scores, respectively, enabling more robust detection of OOD samples. Additionally, we present a semantic-aware content refinement (SaCR) module, which utilizes semantic cues from ID labels to amplify the distributional discrepancy between ID and hard OOD samples. Extensive experiments on several benchmarks demonstrate that \textsc{OT-detector} achieves state-of-the-art performance across various OOD detection tasks, particularly in challenging hard-OOD scenarios.

\end{abstract}

%% file: Sections/1-introduction.tex
\section{Introduction} \label{intro}
Machine learning models are typically trained and evaluated under a closed-set setting, where they are expected to classify data from a predefined set of known classes~\cite{0007LYYL023,TangLZHT25}. However, in real-world scenarios, models are often faced with data from unknown classes~\cite{fu2024cross}, referred to as out-of-distribution (OOD) data. This situation is especially critical in high-stakes applications such as autonomous driving and medical diagnostics, where the failure to detect OOD data may lead to catastrophic outcomes. Therefore, accurately detecting OOD samples is essential for the safe and reliable deployment of machine learning models in practical settings~\cite{DBLP:conf/nips/RenLFSPDDL19,DBLP:conf/nips/TackMJS20,DBLP:conf/nips/XiaoYA20,DBLP:journals/ijcv/YangZLL24,Jiang0GDHL24,jiang2024global,YanXSZT23,YanXTST23}.

\begin{figure}[t!]
    \centering \includegraphics[width=0.45\textwidth]{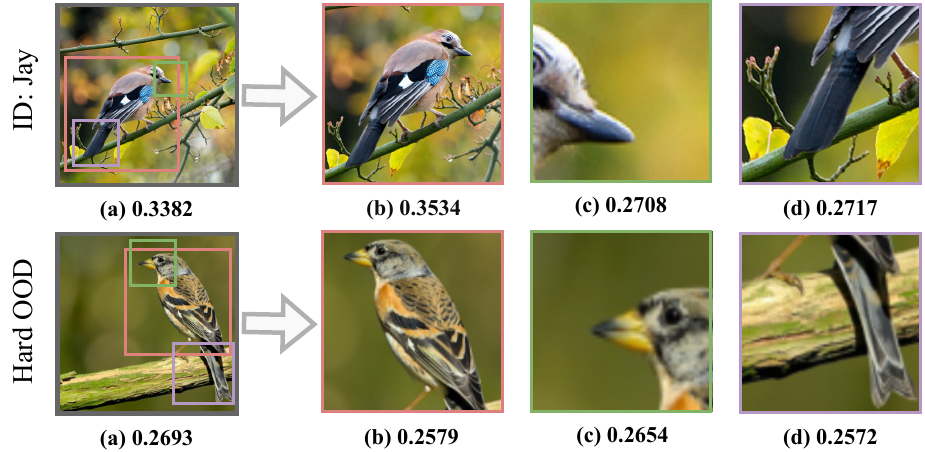}\vspace{-1mm}
    \caption{
    Illustration of diverse views containing distinct semantic information. The value beneath each view represents its cosine similarity to the text prompt ``a photo of a \texttt{Jay}''.
    }
    \label{fig:tsne}\vspace{-2mm}
\end{figure}

Recently, the emergence of Contrastive Language-Image Pre-training (CLIP) has introduced powerful recognition and generalization capabilities~\cite{DBLP:conf/icml/RadfordKHRGASAM21}, thereby enabling various open-world recognition tasks~\cite{shen2024imagpose,shen2025imagdressing}. Building upon CLIP, several \emph{zero-shot} OOD detection methods have been proposed. For example, ZOC~\cite{DBLP:conf/aaai/Esmaeilpour00022} combines visual and textual embeddings to identify OOD samples, and MCM~\cite{DBLP:conf/nips/MingCGSL022} employs a scaled softmax function to match test images against in-distribution (ID) labels. Beyond these approaches, some methods incorporate external knowledge. CLIPN~\cite{DBLP:conf/iccv/WangLYL23} pretrains an additional text encoder with a ``no'' logit to sharpen CLIP’s OOD sensitivity, while NegLabel~\cite{DBLP:conf/iclr/Jiang000LZ024} leverages WordNet to mine negative labels that are semantically dissimilar to ID data. However, these negative labels may not effectively capture OOD samples sharing semantic overlap with ID classes. EOE~\cite{DBLP:conf/icml/CaoZZ0L024} leverages large language models (LLMs) to generate potential OOD labels, but depending on LLMs can introduce privacy and computational concerns. Despite these advances, existing methods predominantly focus on \emph{semantic matching} between test images and ID labels, inherently overlooking \emph{distributional discrepancies} between ID and OOD samples. Moreover, the challenge of distinguishing hard OOD samples that are semantically similar to ID data remains largely unresolved. These gaps raise an important question: \textbf{Can we simultaneously capture both semantic and distributional cues in an unified framework, without resorting to external knowledge?}

To address this issue, we first draw upon the theory of Optimal Transport (OT), a mathematical framework commonly used to quantify discrepancies between probability distributions~\cite{DBLP:journals/pami/CourtyFTR17,DBLP:conf/icml/ArjovskyCB17}. OT has shown promise in settings where known and unlabeled samples coexist~\cite{DBLP:conf/ijcai/RenFXZ24,DBLP:conf/cvpr/Lu0ZZC23,DBLP:journals/pami/CourtyFTR17,DBLP:conf/icml/ArjovskyCB17,DBLP:conf/cvpr/GeLLYS21,DBLP:conf/ijcai/RenFXZ24,DBLP:conf/cvpr/Lu0ZZC23,jiang2024dvf}, yet its application to \emph{zero-shot} OOD detection is non-trivial, particularly due to the lack of ID image samples at inference. Fortunately, CLIP’s vision-text alignment can mitigate this limitation: CLIP embeds relevant text-image pairs closely together in feature space. Motivated by this property, we propose using ID text labels as proxies for ID images. By measuring cosine distances between test images and textual ID labels, OT can naturally identify the closest label for each image. Concretely, we introduce the \emph{transport mass} in OT to represent a \textbf{semantic-wise} score, and a \textbf{distribution-wise} score to measure the \emph{transport cost} bridging the modality gap. By strategically combining these two components, we obtain an OT-based OOD score function that captures both semantic- and distribution-level discrepancies.

However, hard OOD samples that are semantically close to ID classes can still fool CLIP by sharing common backgrounds or object attributes~\cite{Jiang0GDHL24}. As shown in Figure~\ref{fig:tsne}(b), removing ambiguous regions and emphasizing discriminative parts~\cite{TangYLT22} can improve the separability between ID and such OOD samples. Conversely, shared regions (e.g., beaks, tails) lead to low similarity scores unless more fine-grained features are highlighted. To this end, we propose a parameter-free \emph{Semantic-aware Content Refinement} (SaCR) module. For ID images, SaCR refines the visual content by retaining discriminative regions and better aligning features with their corresponding textual labels. For OOD samples, since OOD labels are unavailable, we use ID labels as a guide to select common regions. Consequently, shared regions remain less informative, inducing a feature shift that magnifies the distributional discrepancy between ID and hard OOD in the refined feature space.

By integrating SaCR with the proposed OT-based OOD score function, we develop a new zero-shot OOD detection framework named \textsc{OT-detector}. Experimental results on multiple benchmarks demonstrate that \textsc{OT-detector} achieves robust and reliable performance. In particular, on the ImageNet-1K OOD detection benchmark~\cite{DBLP:conf/nips/HuangGL21}, \textsc{OT-detector} attains a $\text{FPR}{95}$ of $23.65\%$ and $\text{AUROC}$ of $94.49\%$, surpassing existing methods that rely on external knowledge. Notably, \textsc{OT-detector} excels in hard OOD scenarios, setting new state-of-the-art results. Our main contributions are summarized as follows:

\begin{itemize}
\item We introduce \textsc{OT-detector}, a novel zero-shot OOD detection framework that exploits implicit distributional discrepancies via Optimal Transport. To the best of our knowledge, this is the first work applying OT in the zero-shot OOD detection setting.

\item We propose an OT-based OOD score function that captures both semantic-level and distribution-level discrepancies, as well as a parameter-free Semantic-aware Content Refinement module to further separate ID from hard OOD samples.

\item Comprehensive experiments on large-scale and hard OOD benchmarks demonstrate the effectiveness of our approach, establishing new state-of-the-art performance in zero-shot OOD detection.

\end{itemize}

%% file: Sections/2-related.tex
\section{Preliminaries}

\subsection{Contrastive Language-Image Pre-training}
CLIP~\cite{DBLP:conf/icml/RadfordKHRGASAM21} is a multimodal model that leverages a contrastive loss to align textual and visual representations, demonstrating exceptional generalization capabilities. The model consists of a text encoder $\mathcal{T}: u \mapsto \mathbb{R}^d$ and an image encoder $\mathcal{I}: x \mapsto \mathbb{R}^d$, where $u$ is a text sequence and $x$ is an image.
In zero-shot classification tasks, given an input image $x$ and a set of class labels $\mathcal{Y}=\{y_i\}_{i=1}^K$, each label is first embedded into a prompt template (e.g., ``a photo of a [\texttt{CLASS}]''), producing a textual input for the text encoder. Formally:
\begin{equation}
    f^{\text{img}}=\mathcal{I}(x), \quad f^{\text{text}}_i=\mathcal{T}(\operatorname{prompt}(y_i)), 
    \label{eq:extract}
\end{equation}
where $\operatorname{prompt}(\cdot)$ is an operation that inserts the class label into a predefined textual template.
Finally, the predicted label $\hat{y}$ for the image is determined by selecting the class label whose corresponding text feature has the highest cosine similarity with the image feature:
\begin{equation}
    \hat{y}=\underset{y_i \in \mathcal{Y}}{\arg\max}\, \cos\bigl(f^{\text{img}}, f^{\text{text}}_i\bigr).
\label{eq:getlabel}
\end{equation}
This zero-shot inference mechanism allows CLIP to generalize to novel classes not seen during training, relying solely on natural language descriptions.

\subsection{Zero-shot OOD Detection with CLIP}
Zero-shot OOD detection aims to distinguish out-of-distribution (OOD) samples from in-distribution (ID) samples without requiring access to labeled ID training data, leveraging the powerful generalization capabilities of CLIP. The objective is to construct a binary classifier:
\begin{equation}
    G(x;\mathcal{Y}_{\text{id}},\mathcal{I},\mathcal{T}) = \begin{cases} 
    \text{ID}, & S(x) \geq \lambda, \\
    \text{OOD}, & S(x) < \lambda,
    \end{cases}
\end{equation}
where $x$ is an input image sampled from the input space $\mathcal{X}=\mathcal{X}_{\text{id}}\cup\mathcal{X}_{\text{ood}}$, with $\mathcal{X}_{\text{id}}$ and $\mathcal{X}_{\text{ood}}$ denoting sets of ID and OOD samples, respectively. The set $\mathcal{Y}_{\text{id}}$ denotes the ID class labels, and $\lambda$ is a threshold that determines whether a sample is classified as ID or OOD. The OOD score function $S$, as proposed in existing works~\cite{DBLP:conf/nips/MingCGSL022,DBLP:conf/iccv/WangLYL23}, is computed based on the cosine similarity between visual and textual features derived from Eq.~(\ref{eq:extract}).

\subsection{Optimal Transport}
Optimal Transport (OT) focuses on finding the minimum transportation distance required to transfer one distribution to another, such as the Wasserstein distance. This makes it a powerful tool for solving distribution matching problems across various tasks.
Here, we introduce OT for discrete empirical distributions.

Consider two sets, $\mathcal{X}$ with supply samples $\{x_i\}_{i=1}^n$ and $\mathcal{Y}$ with demand samples $\{y_j\}_{j=1}^m$, represented by the empirical distributions:
\begin{equation}
    \mu = \sum_{i=1}^n p_i\delta_{x_i}, \quad \nu=\sum_{j=1}^{m} q_j\delta_{y_j},
\end{equation}
where $\delta_{x_i}$ is the Dirac delta measure centered at $x_i$. The terms $p_i$ and $q_j$ represent the probability masses associated with the samples, satisfying $\sum_{i=1}^n p_{i}=1$ and $\sum_{j=1}^m q_{j}=1$. The minimum transport distance can be determined as follows:
\begin{equation}
    \underset{P\in U(\mu,\nu)}{\min}\langle C,P\rangle_F,
\end{equation}
where 
\begin{equation}
    U(\mu,\nu) = \{P\in \mathbb{R}^{n\times m}_{+} \mid P\mathbf{1}_m = \mu,\; P^\top \mathbf{1}_n=\nu \}
\end{equation}
denotes the transportation polytope of $\mu$ and $\nu$. Here, $\mathbf{1}_n$ and $\mathbf{1}_m$ are all-one vectors of appropriate dimensions. The cost matrix $C\in\mathbb{R}^{n\times m}$ defines the transport cost between supply and demand samples, where each entry corresponds to a chosen metric such as Euclidean or Mahalanobis distance.

To enhance numerical stability and optimization convergence, 
Cuturi~\cite{DBLP:conf/nips/Cuturi13} introduced an entropic regularization term $H(P)$, reformulating the objective as:
\begin{equation}
    \underset{P\in U(\mu,\nu)}{\min}\langle C,P\rangle_F - \frac{1}{\epsilon} H(P),
\end{equation}
where $\epsilon>0$ is the regularization hyperparameter, and
\begin{equation}
    H(P) = -\sum_{i,j} P_{ij}\log P_{ij}.
\end{equation}
The resulting assignment matrix can be expressed as:
\begin{equation} 
P^{\epsilon} = \operatorname{Diag}(\mathbf{a}) \exp(-\epsilon C)\operatorname{Diag}(\mathbf{b}), 
\end{equation}
where $\mathbf{a}$ and $\mathbf{b}$ are non-negative vectors defined up to a multiplicative factor and can be efficiently computed using Sinkhorn’s algorithm~\cite{DBLP:conf/nips/Cuturi13}. 
The OT methodology used in our work is based on the above approach and will be further detailed in the following section.

%% file: Sections/3-method.tex
\section{Method}

\begin{figure}[t!]
    \centering
    \includegraphics[width=0.46\textwidth]{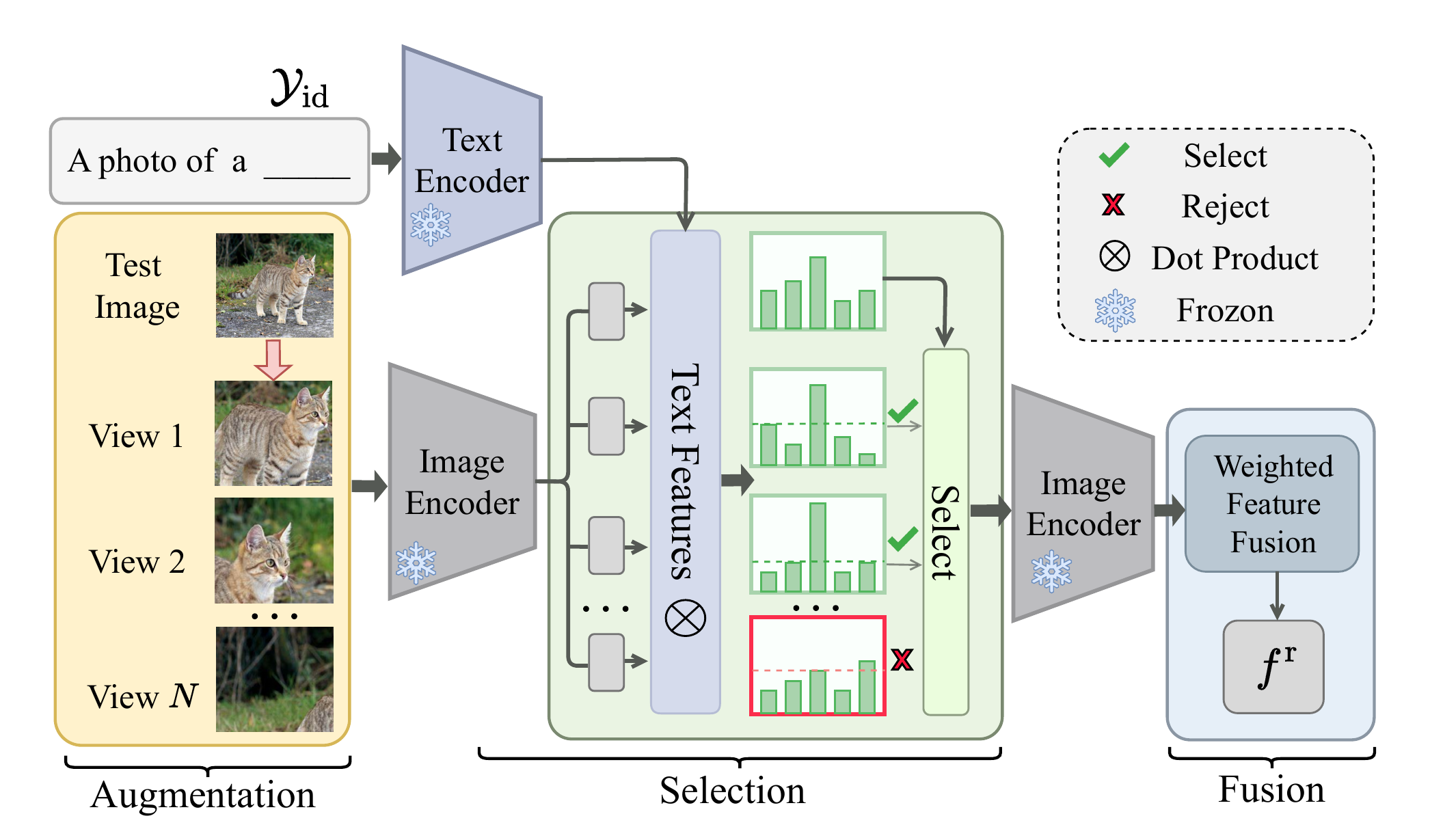}
    \caption{Pipeline of the Semantic-aware Content Refinement (SaCR) module.}\vspace{-2mm}
    \label{fig:cal}
\end{figure}

In this paper, we introduce Optimal Transport (OT) to improve the performance of zero-shot OOD detection. However, two key challenges need to be addressed: 
\begin{enumerate}
\item \textbf{Identifying hard OOD samples} that exhibit high semantic and distributional similarity to ID samples, making them particularly difficult to detect;
\item \textbf{Quantify distributional discrepancies} between ID and OOD samples and transform them into effective OOD detection scores.

\end{enumerate}

To tackle the first challenge, we propose a \emph{Semantic-aware Content Refinement} (SaCR) module, which consists of three steps to adaptively refine both hard OOD and ID samples, as illustrated in Figure~\ref{fig:cal}. By enhancing the distinguishability of refined features, SaCR aims to amplify the distributional discrepancies between ID and OOD samples.

\begin{figure*}[t!]
    \centering
    \includegraphics[width=0.83\textwidth]{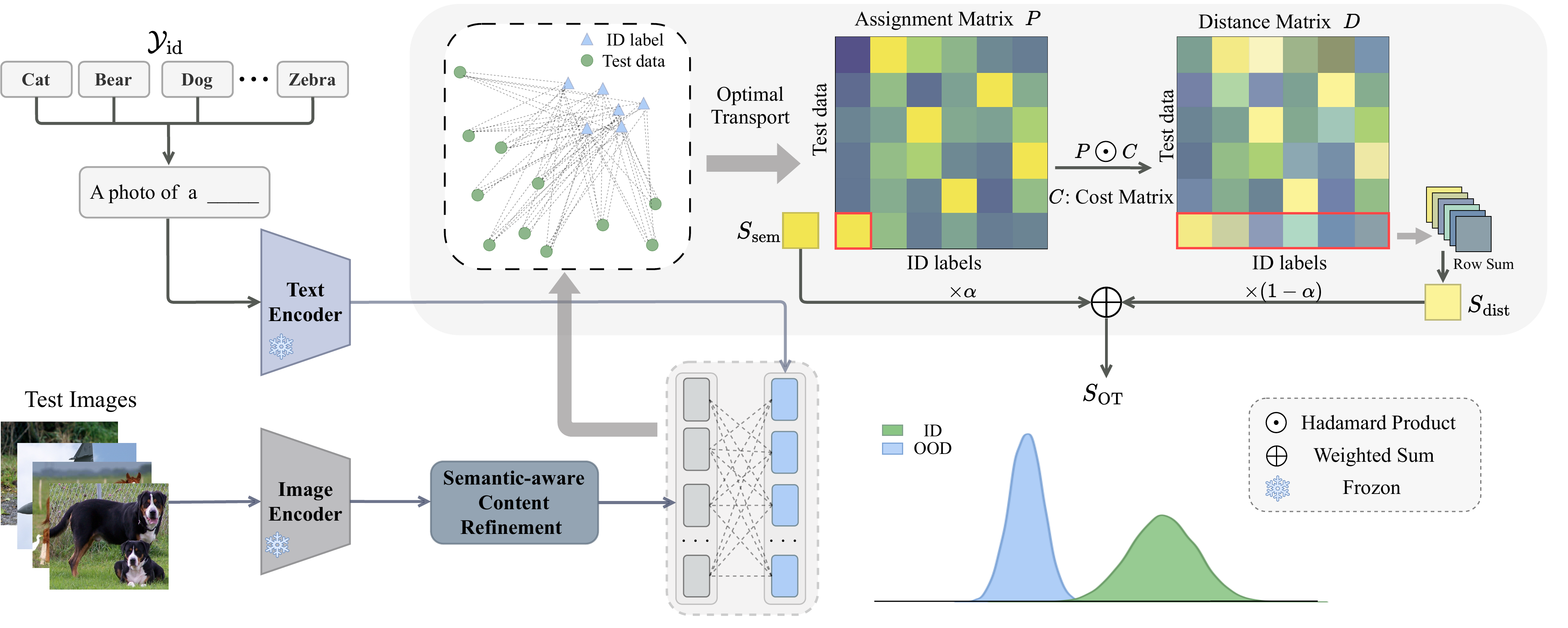}\vspace{-1mm}
    \caption{Pipeline of our Optimal Transport-based framework \textsc{OT-detector} for zero-shot OOD detection. 
    }\vspace{-2mm}
    \label{fig:ot}
\end{figure*}

For the second challenge, we leverage OT to convert the semantic and distributional relationships between test samples and ID label space into an OOD detection score. Specifically, we incorporate two complementary components: a \emph{Semantic-wise OOD Score} and a \emph{Distribution-wise OOD Score}. As shown in Figure~\ref{fig:ot}, we combine these two scores to develop a novel OOD detection framework, termed \textsc{OT-detector}, enabling reliable zero-shot OOD detection without requiring additional ID data or outlier class labels.



\begin{algorithm}[t!]
\caption{Semantic-aware Content Refinement}
\label{alg:FC}
\SetKwInOut{Input}{Input}
\SetKwInOut{Output}{Output}

\newcommand\myCommentStyle[1]{\textcolor{gray!60}{#1}}
\SetCommentSty{myCommentStyle} 

\Input{
    Input image $x$,
    ID labels $\mathcal{Y}_{\text{id}}$,
    Image encoder $\mathcal{I}$,
    Text encoder $\mathcal{T}$,
    Confidence function $M$,
    Top-$k$ selection parameter $k$,
      Number of views $N$
}
\Output{Refined feature $f^\text{r}$}

$X^{\text{v}} \gets \text{RandomCrop}(x, N)$ \tcp{Generate $N$ views}

\For{$y_i \in \mathcal{Y}_{\text{id}}$}{
    $f_i^{\text{text}} \gets \mathcal{T}(\text{prompt}(y_i))$  \tcp{Encode label $y_i$}
}

$\hat{y} \gets \underset{y_i \in \mathcal{Y}_{\text{id}}}{\arg \max} \ \cos\left(\mathcal{I}(x), f_i^{\text{text}}\right)$ \tcp{Predict label $\hat{y}$}

$X^{\text{f}} \gets \emptyset$, $M^{\text{f}} \gets \emptyset$ \\

\For{$x_i^{\text{v}} \in X^{\text{v}}$}{
    $f_i^{\text{img}} \gets \mathcal{I}(x_i^{\text{v}})$ \tcp{Encode view $x_i^{\text{v}}$}
    $\hat{y}_i^{\text{v}} \gets \underset{y_j \in \mathcal{Y}_{\text{id}}}{\arg \max} \ \cos\left(f_i^{\text{img}}, f_j^{\text{text}}\right)$ \tcp{Assign label to view $x_i^{\text{v}}$}
    
    \If{$\hat{y}_i^{\text{v}} = \hat{y}$}{
        $X^{\text{f}} \gets X^{\text{f}} \cup \{x_i^{\text{v}}\}$ \\
        $M^{\text{f}} \gets M^{\text{f}} \cup \{M(x_i^{\text{v}})\}$ \\
    }
}

$inds \gets \text{Top-K}(M^{\text{f}}, k)$ \tcp{Select top-$k$ views}

$f^\text{r} \gets \sum_{i \in inds} M(x_i^{\text{v}}) \cdot \mathcal{I}(x_i^{\text{v}})$ 
\end{algorithm}

\subsection{Semantic-aware Content Refinement} \label{sec:sgfc}
\paragraph{Motivation.}
Hard OOD samples typically exhibit similar background or object attributes to their corresponding ID samples, which confuse CLIP and result in high OOD scores for these hard OOD samples. To mitigate this, we propose a \emph{Semantic-aware Content Refinement} (SaCR) module, which adaptively refines the visual content of both ID and hard OOD samples guided by the ID label semantics, producing more discriminative visual features. By amplifying the distributional discrepancy between ID and hard OOD samples, SaCR makes it easier for the subsequent OT-based score functions to identify hard OOD samples.

\paragraph{View Augmentation.}
We start by generating multiple views of an input image to capture diverse visual information. Specifically, given a test image $x$ and the ID label space $\mathcal{Y}_{\text{id}}$, we apply random cropping at various scales to yield a set of candidate views,
$X^{\text{v}} = \{ x_i^{\text{v}} \}_{i=1}^N$.
This multi-view strategy aims to explore different local regions of $x$ and provide a broader range of features.

\paragraph{View Selection.}
Not all views are equally relevant, some may primarily contain background or only partial object regions. We therefore discard views with predictions inconsistent with that of the original image $x$. Since ground-truth labels are unavailable under the zero-shot setting, we use the semantic concept of $x$ as weak supervision. Let $\hat{y}$ be the predicted label for $x$, and $\hat{Y}^{\text{v}} = \{ y_i^{\text{v}} \}_{i=1}^N$ be the predicted labels for $X^{\text{v}}$. Both $\hat{y}$ and $\hat{Y}^{\text{v}}$ are obtained via Eqs.~(\ref{eq:extract}) and (\ref{eq:getlabel}) with $\mathcal{Y}_{\text{id}}$. We retain only the views whose predicted labels match $\hat{y}$, forming a refined set:
\begin{equation}\label{eq11}
    X^{\text{f}} 
    = \bigl\{ x_i^{\text{v}} \in X^\text{v} \,\big\vert\, \hat{y}_i^{\text{v}} = \hat{y} \bigr\}.
\end{equation}
Next, to further select the most confident views, we define a margin function $M(\cdot)$ based on the cosine similarity logits from CLIP. For an image $x$, let $\mathrm{logits}(x)$ be the sorted similarity scores (logits) over the ID labels. Then the margin $M(x)$ is the difference between the largest and the second-largest logit, i.e.,~$M(x)=\max \mathrm{logits}(x)\;-\;2^\mathrm{nd}\!\max\mathrm{logits}(x)$. We compute $M(x_i^\text{f})$ for each view $x_i^\text{f} \in X^\text{f}$, rank them in descending order, and choose the top-$k$:
\begin{equation}
    X^{\text{f}'} 
    = \Bigl\{ 
        x_j^\text{f} \in X^\text{f} 
        \;\Big\vert\; 
        j \in \operatorname{Top\text{-}k}\bigl(\{ M(x_i^\text{f}) \}_{i=1}^{N'}\bigr)
      \Bigr\}.
\end{equation}
Here, $\operatorname{Top\text{-}k}(\cdot)$ returns the indices of the $k$ highest margin values, and $N' \le N$ is the number of remaining views after the label-consistency filter.

\begin{table*}[t!]
    \centering
    
    \resizebox{0.95\linewidth}{!}{
    \begin{tabular}{lcccccccccc}
        \toprule
        \multirow{3}{*}{\textbf{Method}}  & \multicolumn{10}{c}{\textbf{OOD Dataset}} \\
         ~& \multicolumn{2}{c}{iNaturalist} & \multicolumn{2}{c}{SUN} &\multicolumn{2}{c}{Places } & \multicolumn{2}{c}{Texture} & \multicolumn{2}{c}{\textbf{Average}}\\
        \cmidrule(lr){2-3} \cmidrule(lr){4-5} \cmidrule(lr){6-7} \cmidrule(lr){8-9} \cmidrule(lr){10-11}
        ~& \textbf{FPR95}↓ & \textbf{AUROC}↑ & \textbf{FPR95}↓ &  \textbf{AUROC}↑ & \textbf{FPR95}↓ &  \textbf{AUROC}↑ & \textbf{FPR95}↓ &  \textbf{AUROC}↑ & \textbf{FPR95}↓ &  \textbf{AUROC}↑ \\
        \midrule
        MSP~\cite{DBLP:conf/iclr/HendrycksG17} & 40.89 & 88.63 & 65.81 & 81.24 & 67.90 & 80.14 & 64.96 & 78.16 & 59.89 & 82.04    \\
        Energy~\cite{DBLP:conf/nips/LiuWOL20} & 21.59 & 95.99 & 34.28 & 93.15 & 36.64 & 91.82 & 51.18 & 88.09 & 35.92 & 92.26   \\
        Fort et al.~\cite{DBLP:conf/nips/FortRL21} & 15.07 & 96.64 & 54.12 & 86.37 & 57.99 & 85.24 & 53.32 & 84.77 & 45.12 & 88.25 \\
        CLIPN~\cite{DBLP:conf/iccv/WangLYL23} & 19.13 & 96.20 & 25.69 & 94.18 & 32.14 & 92.26 & 44.60 & 88.93 & 30.39 & 92.89 \\
        MCM~\cite{DBLP:conf/nips/MingCGSL022} & 30.91 & 94.61 & 37.59 & 92.57 & 44.69 & 89.77 & 57.77 & 86.11 & 42.74 & 90.77\\
        GL-MCM~\cite{miyai2023zeroshotindistributiondetectionmultiobject} & 15.18 & 96.71 & 30.42 & 93.09 & 38.85 & 89.90 & 57.93 & 83.63 & 35.47 & 90.83 \\
        EOE~\cite{DBLP:conf/icml/CaoZZ0L024} & 12.29 & 97.52 & 20.40 & 95.73 & 30.16 & 92.95 & 57.53 & 85.64 & 30.09 & 92.96 \\
        NegLabel~\cite{DBLP:conf/iclr/Jiang000LZ024} & \textbf{1.91} & \textbf{99.49} & 20.53 & 95.49 & 35.59 & 91.64 & 43.56 & 90.22 & 25.40 & 94.21 \\
        \rowcolor{gray! 20} \textsc{OT-detector}~(Ours) & 7.91 & 98.16 & \textbf{16.18} & \textbf{96.15} & \textbf{27.43} & \textbf{93.30} & \textbf{43.07} & \textbf{90.34} & \textbf{23.65} & \textbf{94.49} \\
        \bottomrule
    \end{tabular}
    }
    \caption{Performance comparison for ImageNet-1K benchmark as ID dataset with various OOD datasets.}\vspace{-1mm}
    \label{Image1K}
\end{table*}

\paragraph{View Fusion.}
Finally, we fuse the selected views into a single, refined feature. Each view is first encoded by the CLIP image encoder $\mathcal{I}(\cdot)$, then weighted by its margin score and summed:
\begin{equation}
    f^{\text{r}} 
    = \operatorname{norm}\Bigl(\sum_{j=1}^{k} M(x_j^{\text{f}}) \cdot \mathcal{I}\bigl(x_j^{\text{f}}\bigr)\Bigr),
\end{equation}
where $\operatorname{norm}(\cdot)$ denotes L2 normalization along the feature dimension. A step-by-step summary of this procedure is outlined in Algorithm~\ref{alg:FC}.

\subsection{A New OOD Detection Score}\label{sec:ot_score}
\textbf{Motivation.}
Empirically, OOD and ID data often occupy distinct regions in the embedding space. This observation motivates us to leverage Optimal Transport (OT) to quantify this distributional discrepancy and design an effective OOD score. Since direct interaction between ID and OOD samples is not feasible, we employ ID label features as a bridge and use OT to measure the distributional difference between test samples and these label features. Specifically, we propose two complementary OT-based scores in our \textsc{OT-detector}, which consider both sample-level and distribution-level discrepancies:
\begin{itemize}
\item \textbf{Semantic-wise OOD Score:} 
OT assigns larger transport mass to supply samples that are closer to demand samples. Hence, the mass transported from a test sample to an ID label can reflect their semantic similarity. For OOD samples that tend to be far from all ID labels, this transported mass will be relatively small.
\item  \textbf{Distribution-wise OOD Score:} 
Considering the modality gap between textual and visual features, we further identify OOD samples by assessing the cost required for a test sample to ``bridge'' this gap to the ID label feature space. This cost, derived from the OT distance, indicates the distributional discrepancy.
\end{itemize}

\paragraph{OT Formulation.}
Let the test set be $\mathcal{X} = \{ x_i \}_{i=1}^{|\mathcal{X}|}$ and the refined visual features (from the Semantic-aware Content Refinement module in Section~\ref{sec:sgfc}) be
$
F^{\text{r}} = \{ f_i^{\text{r}} \}_{i=1}^{|\mathcal{X}|} \in \mathbb{R}^{|\mathcal{X}|\times d},
$
where $|\mathcal{X}|$ denotes the number of test samples, and $d$ is the feature dimension.
Meanwhile, we denote the CLIP textual features for all ID labels by
$
F^{\text{text}} = \{ f_j^{\text{text}} \}_{j=1}^{K} \in \mathbb{R}^{K\times d},
$
where $K$ is the number of ID class labels.
We assume unknown priors and define the discrete measures:
\begin{equation}
\mu = \sum_{i=1}^{|\mathcal{X}|} p_i \,\delta_{f_i^{\text{r}}}, 
\quad 
\nu = \sum_{j=1}^{K} q_j \,\delta_{f_j^{\text{text}}},
\end{equation}
with $p_i = \frac{1}{|\mathcal{X}|}$ and $q_j = \frac{1}{K}$. 
To construct the cost matrix $C \in \mathbb{R}^{|\mathcal{X}|\times K}$, we adopt the cosine distance since CLIP is pretrained with a contrastive objective that aligns visual and textual representations, i.e.,~$C = 1 - F^\text{r} \,\bigl(F^{\text{text}}\bigr)^{\top}$.
We then solve the following entropic regularized OT problem:
\begin{equation}\label{eq14}
    \min_{P\in U(\mu,\nu)} 
    \Bigl\langle 
        1 - F^\text{r}\,\bigl(F^\text{text}\bigr)^\top, 
        P
    \Bigr\rangle_F 
    \;-\; \frac{1}{\epsilon} \, H(P),
\end{equation}
where $H(P)$ is the entropic regularization term, $\epsilon > 0$ is the regularization parameter, and $P^* \in \mathbb{R}^{|\mathcal{X}|\times K}$ denotes the optimal transport plan.

\paragraph{Semantic-wise OOD Score.}
The assignment matrix $P^*$ maps $|\mathcal{X}|$ test samples to $K$ ID labels, capturing the transport mass from each test sample $x_i$ to every ID label. We quantify the semantic alignment by focusing on the largest transport mass for each sample. Formally, the semantic-wise score is 
\begin{equation}
    S_\text{sem}(x_i) = \max_{j} \, p^*_{ij}.
\end{equation}
Intuitively, higher mass indicates stronger alignment with at least one ID label.

\paragraph{Distribution-wise OOD Score.}
The Wasserstein distance, $\sum_{i,j} p^*_{ij} \, c_{ij}$, reflects the overall distributional discrepancy between test samples and ID labels. We decompose it to isolate the portion that corresponds to each test sample $x_i$, defining the distribution-wise score as
\begin{equation}
    S_\text{dist}(x_i) = 1 - \sum_{j=1}^K p^*_{ij}\,c_{ij}.
\end{equation}
Here, $c_{ij}$ is the entry of the cost matrix $C$, and for OOD samples, bridging the gap to the ID label space typically incurs a larger cost, leading to a lower $S_\text{dist}$.

\paragraph{Final OT-based OOD Score Function.}
We combine these two score functions in our \textsc{OT-detector} to effectively capture both semantic alignment and distributional discrepancy:
\begin{equation}
    S_{\text{OT}}(x_i) 
    = \alpha \, S_{\text{sem}}(x_i)
    + (1 - \alpha)\, S_{\text{dist}}(x_i),
    \label{eq:S_ot}
\end{equation}
where $\alpha \in [0,1]$ balances the contributions of $S_{\text{sem}}$ and $S_{\text{dist}}$.

    


%% file: Sections/4-experiments.tex
\section{Experiments}
\subsection{Experiments Setup}
\paragraph{Datasets.} Following the previous work~\cite{DBLP:conf/nips/MingCGSL022}, we evaluate our method on the ImageNet-1K OOD benchmark to ensure comparability with other methods. The ImageNet-1K OOD benchmark uses ImageNet-1K as the ID data and considers Texture ~\cite{DBLP:conf/cvpr/CimpoiMKMV14}, iNaturalist ~\cite{DBLP:conf/cvpr/HornASCSSAPB18}, SUN ~\cite{DBLP:conf/cvpr/XiaoHEOT10}, and Places365 ~\cite{DBLP:journals/pami/ZhouLKO018} as OOD data.
Additionally, we perform hard OOD analysis using semantically similar subsets of ImageNet constructed by MCM, i.e.,~ImageNet-10, ImageNet-20, and ImageNet-100, which show high semantic similarity.

\begin{table*}[t!]
    \centering
    \resizebox{0.95\linewidth}{!}{
    \begin{tabular}{lccccccccccc}
        \toprule
         \multirow{3}{*}{\textbf{Method}} & \textbf{ID} & \multicolumn{2}{c}{ImageNet-10}& \multicolumn{2}{c}{ImageNet-20} &  \multicolumn{2}{c}{ImageNet-10} & \multicolumn{2}{c}{ImageNet-100} &\multicolumn{2}{c}{\multirow{2}{*}{\textbf{Average}}}\\
         & \textbf{OOD} & \multicolumn{2}{c}{ImageNet-20}& \multicolumn{2}{c}{ImageNet-10} &  \multicolumn{2}{c}{ImageNet-100} & \multicolumn{2}{c}{ImageNet-10}\\
        \cmidrule(lr){3-4} \cmidrule(lr){5-6} \cmidrule(lr){7-8} \cmidrule(lr){9-10} \cmidrule(lr){11-12} 
         && \textbf{FPR95}↓ & \textbf{AUROC}↑ & \textbf{FPR95}↓ &  \textbf{AUROC}↑ & \textbf{FPR95}↓ &  \textbf{AUROC}↑ & \textbf{FPR95}↓ &  \textbf{AUROC}↑& \textbf{FPR95}↓ &  \textbf{AUROC}↑  \\
        \midrule
        Energy~\cite{DBLP:conf/nips/LiuWOL20} & & 10.20 & 97.94 & 18.88 & 97.15 & 5.65 & 98.90 & 63.51 &87.66 & 24.56 & 95.41 \\
        MCM~\cite{DBLP:conf/nips/MingCGSL022}&&5.00& 98.71 & 16.99& 97.69& 2.29 & 99.31 & 66.37 & 86.45 & 22.66 & 95.54\\
        EOE~\cite{DBLP:conf/icml/CaoZZ0L024}&& 4.20 & 99.09 &15.85 &97.74 & 5.13 & 98.85 & 68.71 & 85.62 & 23.47 & 95.33\\
        NegLabel~\cite{DBLP:conf/iclr/Jiang000LZ024}& &5.10& 98.86 & 4.60 & 98.81& \textbf{1.68} & \textbf{99.51} & 40.20 & 90.19 & 12.90 & 96.84 \\
        \rowcolor{gray! 20} \textsc{OT-detector}~(Ours) &&\textbf{2.50} & \textbf{99.17}& \textbf{3.78} & \textbf{99.14} & 3.06& 99.17 & \textbf{22.60} & \textbf{96.15} & \textbf{7.98}& \textbf{98.41}\\
        \bottomrule
    \end{tabular}
    }\vspace{-1mm}
    \caption{Performance comparison on hard OOD detection tasks.}\vspace{-1mm}
    \label{hardood}
\end{table*}

\paragraph{Evaluation Metrics.} We evaluate our method using two standard metrics:(1) \textbf{FPR95:} the false positive rate of OOD samples when the true positive rate of ID samples is at $95\%$; (2) \textbf{AUROC:} the area under the receiver operating characteristic curve.

\paragraph{Implementation Details.} We utilize CLIP as the pretrained model, widely adopted in previous works. Specifically, we employ CLIP-B/16, comprising a ViT-B/16 image encoder and a masked self-attention Transformer text encoder, with weights from OpenAI's open-source models. 
For view augmentation, $N = 256$ randomly cropped images are generated, and the $\text{Top-}k=20$ crop features are selected for fusion. In the OT component, we fix $ \epsilon = 90$ for entropic regularization and dynamically determine the optimal $ \alpha $ for each OOD dataset.
We also analyze the impact of different batch sizes (see Appendix~\ref{ap:limit}). 
Since this work does not focus on textual modality, we use a standard prompt, "a photo of a [\texttt{CLASS}]," for all experiments. 
Further analysis on the influence of different prompts is provided in Appendix~\ref{ap:prompt}. 

\paragraph{Compared Methods.} We compare our method with state-of-the-art OOD detection methods, including zero-shot methods and those requiring fine-tuning or auxiliary OOD labels. For a fair comparison, all methods are implemented using CLIP-B/16 as their backbone, consistent with our \textsc{OT-detector}. For methods requiring pre-training or fine-tuning, we condider MSP~\cite{DBLP:conf/iclr/HendrycksG17}, Energy~\cite{DBLP:conf/nips/LiuWOL20}, CLIPN~\cite{DBLP:conf/iccv/WangLYL23}, and the method proposed in~\cite{DBLP:conf/nips/FortRL21}. Among zero-shot methods, we compare against MCM~\cite{DBLP:conf/nips/MingCGSL022} and GL-MCM~\cite{miyai2023zeroshotindistributiondetectionmultiobject}. 
Additionally, we evaluate our \textsc{OT-detector} against recently proposed representative methods, including EOE~\cite{DBLP:conf/icml/CaoZZ0L024} and NegLabel~\cite{DBLP:conf/iclr/Jiang000LZ024}. Notably, CLIPN
relies on a large-scale auxiliary dataset, NegLabel leverages a large corpus database, and EOE depends on large language models. In contrast, our \textsc{OT-detector} operates without requiring any auxiliary datasets or candidate OOD labels.

\subsection{Main Results}
\label{mainresult}
\paragraph{Evaluation on ImageNet-1K Benchmark.} Table~\ref{Image1K} shows the performance of our method on the large-scale ImageNet-1K dataset as ID data, evaluated against four OOD datasets: iNaturalist, SUN, Places, and Texture. Our method achieves state-of-the-art (SOTA) performance, with an average FPR95 of $23.65\%$ and AUROC of $94.49\%$. Notably, it significantly outperforms recently proposed EOE and NegLabel.
Further discussion and analysis are provided in Appendix~\ref{ap:neglabel}.

\paragraph{Evaluation on Hard OOD Detection Tasks.}
Table~\ref{hardood} presents the experimental result of our method on hard OOD detection tasks. \textsc{OT-detector} consistently demonstrates superior results across all four tasks. Specifically, when ImageNet-100 is used as ID data and ImageNet-10 as OOD data, our method achieves $43.78\%$ improvement in FPR95 and $6.61\%$ improvement in AUROC compared to NegLabel, without requiring access to potential outlier OOD labels. Finally, our method achieves an average FPR95 of $7.98\%$ and AUROC of $98.41\%$ on hard OOD tasks. These results highlight the outstanding effectiveness of \textsc{OT-detector} in tackling challenging OOD detection scenarios.

    



\begin{table}[t!]
    \centering
    
    \resizebox{0.7\linewidth}{!}{
    \begin{tabular}{ccccc}
        \toprule
        \multirow{2}{*}{SaCR} & \multicolumn{2}{c}{Score Functions} & \multirow{2}{*}{\textbf{FPR95}↓} & \multirow{2}{*}{\textbf{AUROC}↑} \\
        &$S_{\text{sem}}$&$S_{\text{dist}}$& &\\
        \midrule
        \ding{55} & \ding{55} & \ding{55}& 42.74 & 90.77\\
        \midrule

        \ding{55} & \ding{51} & \ding{55}& 31.30 & 92.07\\
        \ding{55} & \ding{55} & \ding{51}& 43.76 & 90.94\\
        \ding{55} & \ding{51} & \ding{51}& 29.54 &92.92\\
        \midrule
        
        \ding{51} & \ding{55} & \ding{55}& 38.67 & 91.31\\
        \ding{51} & \ding{51} & \ding{55}& 27.88 &  93.28\\
        \ding{51} & \ding{55} & \ding{51}& 29.25& 93.74\\

        \ding{51} & \ding{51} & \ding{51}& \textbf{23.65} & \textbf{94.49}\\
        \bottomrule
    \end{tabular}
    }
    \caption{Ablation study of main components on the ImageNet-1K.}\vspace{-2mm}
    \label{abaltion}
\end{table}

\subsection{Ablation Study} \label{abaltionstudy}
Table~\ref{abaltion} analyzes the key components of \textsc{OT-detector}. Initially, MCM~\cite{DBLP:conf/nips/MingCGSL022} serves as our baseline when no additional modules are applied.

\paragraph{OT-based Score Functions.}
Comparing the first and second rows of Table~\ref{abaltion}, we observe that using only the semantic-wise score $S_\text{sem}$ notably outperforms the baseline, indicating that the transport mass from the assignment matrix effectively captures semantic discrepancy crucial for identifying OOD samples. In contrast, comparing the first and third rows shows that the distribution-wise score $S_\text{dist}$ alone achieves an FPR95 of $43.76\%$, performing on par with the baseline. This suggests that $S_\text{dist}$ primarily captures global distributional discrepancy. However, combining $S_\text{dist}$ with $S_\text{sem}$ fully leverages their complementary strengths, resulting in a $30.88\%$ improvement in FPR95 and a $2.37\%$ increase in AUROC over the baseline.

\paragraph{Semantic-aware Content Refinement.}
Comparing the first and fifth rows, as well as the second and fifth rows, shows that SaCR provides moderate improvements when using semantic-based scores. More notably, comparing the third and seventh rows highlights a substantial enhancement in $S_{\text{dist}}$ with SaCR, yielding a $33.16\%$ reduction in FPR95 and a $2.99\%$ increase in AUROC. This finding indicates that SaCR introduces distributional shifts that make ID and OOD samples more distinguishable, thereby enabling our OOD score to more effectively capture these differences and enhancing the discriminative power of $S_{\text{dist}}$. Finally, comparing the fourth and last rows shows that applying SaCR to the OT-based OOD score yields a further $19.94\%$ reduction in FPR95 and a $1.66\%$ increase in AUROC. These results confirm the effectiveness of SaCR in our \textsc{OT-detector}, significantly boosting OOD detection performance.


\begin{figure}[t!]
    \centering
    \includegraphics[width=0.45\textwidth]{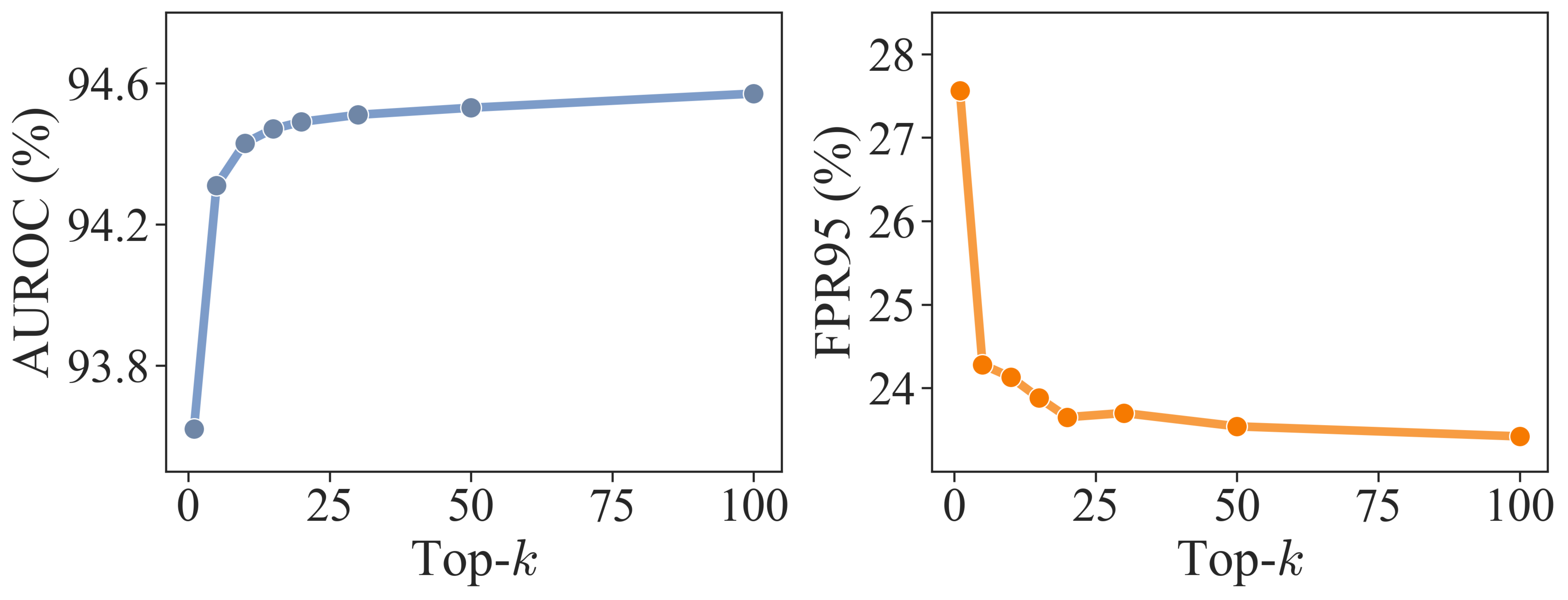}\vspace{-2mm}
    \caption{Analyses on the hyper-parameter of threshold $k$, where results are reported with
ImageNet-1K benchmark.}\vspace{-1mm}
    \label{fig:k}
\end{figure}

\subsection{Sensitivity Analysis}

\paragraph{Top-$k$ in Eq.~\ref{eq11}.}
Figure~\ref{fig:k} shows the effect of varying Top-$k$ (i.e.,~$k \in \{1, 5, 10, 15, 20, 30, 50, 100\}$) on the performance of our SaCR module. 
After filtering views for semantic consistency, SaCR selects the top-$k$ views based on their confidence scores for fusion. 
Since higher-confidence views typically contain more informative content, they are prioritized for feature refinement. 
As presented in Figure~\ref{fig:k}, the largest performance gain occurs at $k=100$.
But increasing $k$ introduces additional views of lower confidence, resulting in only marginal changes to the fused features. To balance between capturing sufficiently informative views and avoiding redundancy, we set $k=20$ by default.


\begin{table}[t!]
    \centering
    
    \resizebox{\linewidth}{!}{
    \begin{tabular}{ccccccccc}
        \toprule
         \multirow{2}{*}{$\alpha$ }& \multicolumn{2}{c}{iNaturalist} &\multicolumn{2}{c}{SUN} & \multicolumn{2}{c}{Places} & \multicolumn{2}{c}{Texture}\\
        \cmidrule(lr){2-3} \cmidrule(lr){4-5} \cmidrule(lr){6-7} \cmidrule(lr){8-9} 
        ~& \textbf{FPR95}↓ & \textbf{AUROC}↑ & \textbf{FPR95}↓ &  \textbf{AUROC}↑ & \textbf{FPR95}↓ &  \textbf{AUROC}↑ & \textbf{FPR95}↓ &  \textbf{AUROC}↑  \\

        \midrule
        0 & 10.45 & 97.68 & 18.45 & 95.94& \cellcolor{HighLight}27.43& \cellcolor{HighLight}93.30&   60.67& 88.06 \\
        0.1 & 8.65 & 97.95 & 16.65 & 96.11& 27.37& 93.21&   54.40& 89.12 \\
        0.2 & 8.10 & 98.09 & \cellcolor{HighLight}16.18 & \cellcolor{HighLight}96.15& 27.44& 92.98&   49.27& 89.75 \\
        0.3 & \cellcolor{HighLight}7.91 & \cellcolor{HighLight}98.16 & 16.47 & 96.11& 28.28& 92.68&   46.81& 90.09 \\
        0.4 & 8.05 & 98.18 & 17.00 & 96.01& 29.22& 92.35&   44.73& 90.27 \\
        0.5 & 8.22 & 98.17 & 17.83 & 95.89& 30.31& 92.00&   43.95& 90.34 \\
        0.6 & 8.53 & 98.13 & 18.63 & 95.74& 31.28& 91.65&  \cellcolor{HighLight} 43.07& \cellcolor{HighLight}90.34 \\
        0.7 & 8.84 & 98.08 & 19.35 & 95.57& 32.60& 91.31&   42.85& 90.30 \\
        0.8 & 9.25 & 98.01 & 20.23 & 95.38& 33.75& 90.97&   42.77& 90.23 \\
        0.9 & 9.79 & 97.92 & 21.20 & 95.19& 34.74& 90.63&   43.05& 90.14 \\
        1 & 10.44 & 97.82 & 22.07 & 94.98& 35.83& 90.30&   43.19& 90.03 \\
        \bottomrule
    \end{tabular}
    }
    \caption{Effect of weight $\alpha$ for each score in Eq. (\ref{eq:S_ot}).The ID dataset is ImageNet-1K. $\alpha$ used for different OOD datasets are highlighted.}
    \label{abaltion:alpha}\vspace{-1mm}
\end{table}

\paragraph{Hyper-parameter $\alpha$ in Eq.~\ref{eq:S_ot}.} Table~\ref{abaltion:alpha} provides a detailed analysis of $\alpha$ values ranging from $0$ to $1$ in increments of $0.1$. 
We can see that incorporating both scores with a proper balance consistently yields better performance, as dataset statistics can vary substantially and thus favor different scoring components. For the SUN and Places datasets, the distribution-wise score ($\alpha=0$) surpasses the semantic-wise score ($\alpha=1$). In contrast, on the Texture dataset, using only the distribution-wise score performs poorly. Introducing the semantic-wise component ($\alpha>0$) notably improves FPR95 by $29.01\%$, suggesting that the large semantic gap between texture images and ImageNet-1K classes benefits significantly from semantic alignment. These observations demonstrate that our score function can adapt effectively to various dataset characteristics by appropriately tuning $\alpha$.

\subsection{Visualization}

\begin{figure}[t!] 
    \centering
    \includegraphics[width=0.45\textwidth]{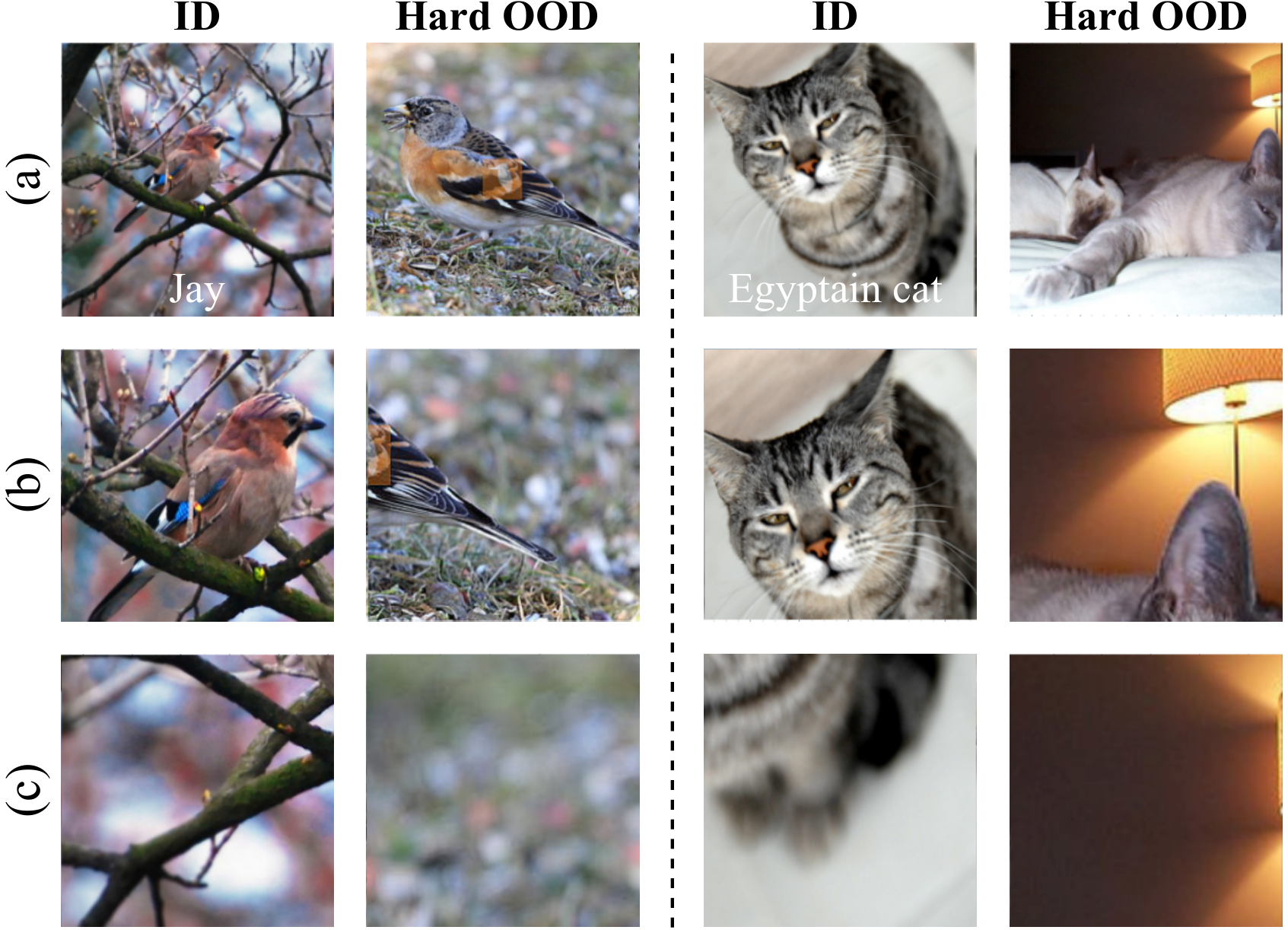}
    \caption{Visualization of SaCR for ID/Hard OOD sample pairs: (a) Test Images; (b) Views selected with largest margin; (c) Views filtered with lower margin.}
    \label{fig:vis}\vspace{-2mm}
\end{figure}

\paragraph{Case Visualization.} 
Figure~\ref{fig:vis} illustrates two pairs of ID and hard OOD samples, with ImageNet-100 as the ID dataset and ImageNet-10 as the OOD dataset. Specifically, Figure~\ref{fig:vis}(a) presents the original test images, Figure~\ref{fig:vis}(b) shows the views with the highest confidence selected by SaCR, and Figure~\ref{fig:vis}(c) highlights views with low margins due to insufficient semantic information. For the ID image, SaCR effectively selects views that emphasize the object while eliminating irrelevant background details. In contrast, the hard OOD image in the second column of Figure~\ref{fig:vis}(a) is initially misclassified by CLIP as the ``\texttt{jay}'' class, and then SaCR selects views focusing on the object parts most resembling the ``\texttt{jay}'' concept, specifically the bird's tail, as depicted in the second image of Figure~\ref{fig:vis}(b). As a result, the fused features of the ID and hard OOD samples become significantly more distinct, enhancing their distributional discrepancy. Through this refinement, SaCR automatically identifies relevant parts of the hard OOD samples, thereby amplifying the gap between ID and hard OOD samples. This enhanced separation is then captured by the OT-based OOD score function, which leads to improved detection of hard OOD samples.

\begin{figure}[t!]
    \centering
    \begin{subfigure}[b]{0.16\textwidth}
        \centering
        \includegraphics[width=\textwidth]{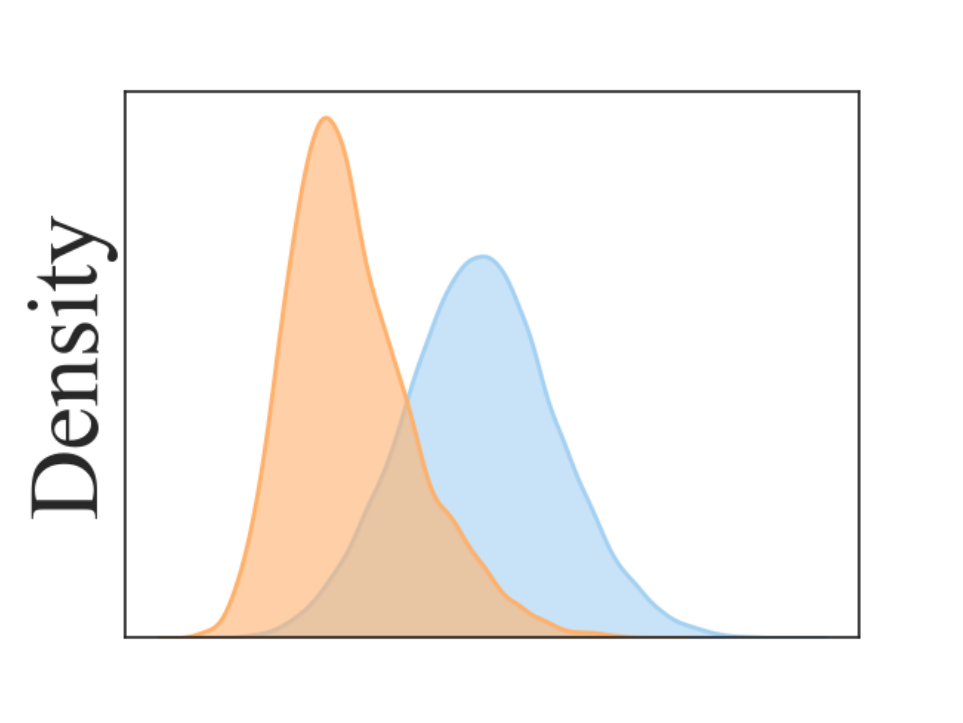}\vspace{-2mm}
        \caption{MCM}
    \end{subfigure}
    \hspace{-4mm}
    \begin{subfigure}[b]{0.16\textwidth}
        \centering
        \includegraphics[width=\textwidth]{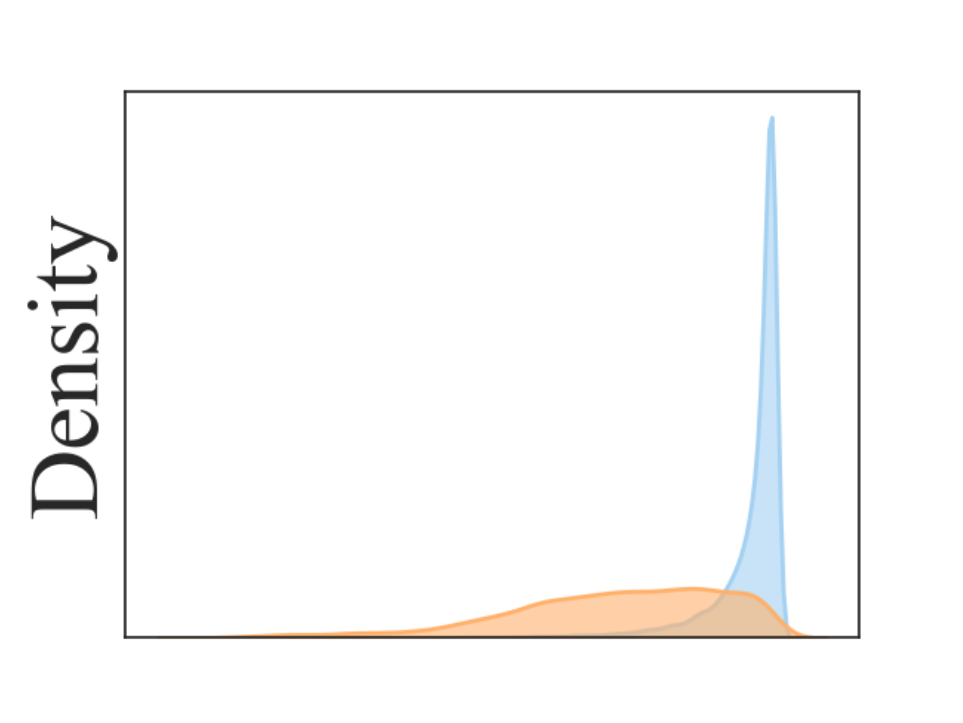}\vspace{-2mm}
        \caption{NegLabel}
    \end{subfigure}
    \hspace{-4mm}
     \begin{subfigure}[b]{0.16\textwidth}
        \centering
        \includegraphics[width=\textwidth]{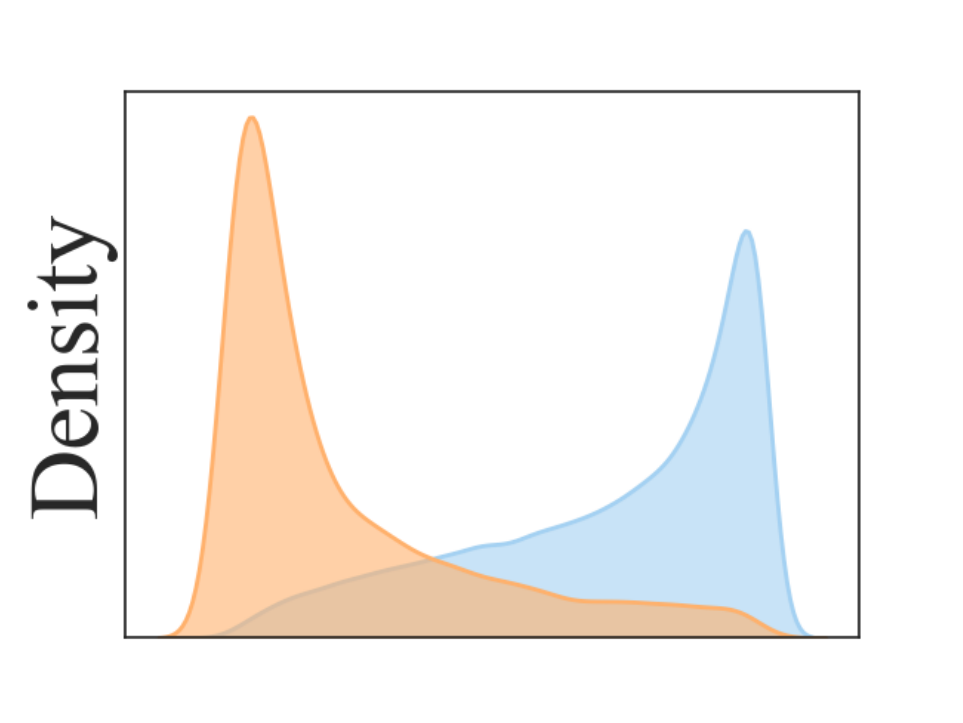}\vspace{-2mm}
        \caption{\textsc{OT-detector}}
    \end{subfigure}\vspace{-2mm}
    \caption{Score distribution of the ID/OOD score with ImageNet-1K/Places as ID/OOD data.}
    \label{fig:density}\vspace{-1mm}
\end{figure}

\paragraph{Score Distribution.} 
Figure~\ref{fig:density} presents the density plots of ID and OOD scores for MCM~\cite{DBLP:conf/nips/MingCGSL022}, NegLabel~\cite{DBLP:conf/iclr/Jiang000LZ024}, and \textsc{OT-detector}, using the ImageNet-1K dataset as ID data and the Places dataset as OOD data. In Figure~\ref{fig:density}(a), MCM exhibits a bimodal distribution for both ID and OOD scores. However, a significant overlap exists between the two distributions, making it difficult to determine a suitable threshold for effectively distinguishing OOD data from ID data. Figure~\ref{fig:density}(b) shows the score distribution for NegLabel. The introduction of numerous negative labels causes ID scores to cluster at higher values due to the negative mining process, which relies on words that are semantically dissimilar to ID labels. However, when OOD images share certain semantic similarity with ID data, the mined negative labels fail to match well, resulting in high confidence scores for many OOD samples. In contrast, Figure~\ref{fig:density}(c) illustrates the density plot for \textsc{OT-detector}, revealing a clear bimodal distribution with substantially reduced overlap between ID and OOD scores relative to MCM. This distinct separation of distributions enables more effective threshold selection, thereby leading to superior OOD detection performance by \textsc{OT-detector}.


%% file: Sections/5-conclusion.tex
\section{Conclusion}
In this paper, we presented a novel and effective framework for zero-shot OOD detection, termed \textsc{OT-detector}, by incorporating Optimal Transport to quantify semantic and distributional discrepancies. Building upon this foundation, we devised an OT-based OOD score function that integrates both semantic and distributional scores. Furthermore, to tackle the challenge of hard OOD detection without relying on external knowledge, we introduced a semantic-aware content refinement module, which adaptively amplifies the distinction between ID and hard OOD samples. Extensive experiments demonstrate that \textsc{OT-detector} achieves state-of-the-art results across various zero-shot detection benchmarks and excels in detecting hard OOD samples.

%% file: Sections/Appendix.tex
\appendix
\section{Appendix}

This appendix provides supplementary materials and detailed discussions to support the main content of the paper. Specifically:

\begin{itemize}
    \item \textbf{Section~\ref{ap:related}} offers a comprehensive review of related works.
    \item \textbf{Section~\ref{ap:prompt}} presents extensive evaluations of our \textsc{OT-detector} using various prompt templates.
    \item \textbf{Section~\ref{ap:various_ID}} reports experimental results of our \textsc{OT-detector} on additional tasks.
    \item \textbf{Sections~\ref{ap:ot} and~\ref{ap:sacr}} contain further experimental results on the components of our \textsc{OT-detector}.
    \item \textbf{Section~\ref{ap:backbone}} validates the effectiveness of our \textsc{OT-detector} using different versions of CLIP.
    \item \textbf{Section~\ref{ap:limit}} discusses the limitations of our \textsc{OT-detector} and proposes potential future directions.
    \item \textbf{Section~\ref{ap:neglabel}} provides a comparative analysis of the strengths and weaknesses of NegLabel in relation to our \textsc{OT-detector}.
\end{itemize}

\subsection{Related Works}\label{ap:related}
\paragraph{OOD Detection.} 
OOD detection aims to identify inputs that deviate semantically from the training data. This task is typically categorized into three main approaches: score-based methods~\cite{DBLP:conf/iccv/YangWFYZZ021,DBLP:conf/icml/MingFL22,DBLP:conf/cvpr/Wang0F022}, distance-based approaches~\cite{DBLP:conf/nips/TackMJS20,DBLP:conf/icml/SunM0L22,DBLP:conf/nips/DuGML22,DBLP:conf/aaai/GhosalSL24}, and generative-based approaches~\cite{DBLP:conf/nips/XiaoYA20,DBLP:conf/nips/RenLFSPDDL19,DBLP:conf/iccv/KongR21,DBLP:conf/cvpr/ChauhanUSGS22}. 
Score-based methods typically design a score function based on the model's uncertainty for a given test input, which is then used to determine whether the input is OOD. For instance, \cite{DBLP:conf/iclr/HendrycksG17} introduce the maximum softmax probability as a baseline for OOD detection. Additionally, \cite{DBLP:conf/iclr/LiangLS18} observe that adversarial perturbations during training can improve OOD detection, while \cite{DBLP:conf/nips/LiuWOL20} propose the energy score, which aligns with the probability density for OOD detection. ReAct~\cite{DBLP:conf/nips/SunGL21} improves OOD detection by dynamically adjusting confidence thresholds and incorporating rejection regions. Mahalanobis distance-based method~\cite{DBLP:conf/nips/LeeLLS18} leverages class-conditional feature distributions to measure the likelihood of a sample being OOD. More recently, \cite{DBLP:conf/icml/LiuZWW23} utilize the manifold mapping capability of diffusion models, employing a standard image similarity metric to quantify the distance between original and reconstructed images for OOD detection.
However, these methods often rely solely on the visual modality and exhibit suboptimal performance when applied to large-scale benchmarks. To address this, recent advances in zero-shot OOD detection that integrate Vision-Language Models (VLM) with OOD detection have shown promising results.

\paragraph{Zero-shot OOD Detection.}
Zero-shot OOD detection aims to identify samples that do not belong to any ID class described by user-provided text, leveraging the semantic understanding capabilities of vision-language models (VLMs). Most existing methods identify OOD samples based on the cosine similarity between visual and textual features. For example, MCM~\cite{DBLP:conf/nips/MingCGSL022} uses the maximum temperature-scaled softmax value between the textual features of ID labels and the visual features of test images as the OOD score. Building on MCM, GL-MCM~\cite{miyai2023zeroshotindistributiondetectionmultiobject} incorporates background information, designing a global OOD score that also accounts for the similarity between background features and textual features. ZOC~\cite{DBLP:conf/aaai/Esmaeilpour00022} introduces an additional text decoder as a captioner to generate candidate OOD labels for test data. Similarly, CLIPN~\cite{DBLP:conf/iccv/WangLYL23} trains an additional text encoder to enable CLIP to effectively handle negative prompts. EOE~\cite{DBLP:conf/icml/CaoZZ0L024} leverages large language models (LLMs) by designing prompts to guide LLMs in generating candidate OOD labels, improving OOD detection performance. However, the labels generated through LLMs lack sufficient diversity, which consequently reduces the performance of OOD detection. NegLabel~\cite{DBLP:conf/iclr/Jiang000LZ024} introduces a NegMining algorithm to extract candidate OOD labels by selecting semantically distant words from WordNet. It designs an OOD score based on the similarity to these negative labels. However, since OOD samples may exhibit a certain degree of semantic similarity to ID samples, these negative labels often struggle to align with such OOD samples, leading to limited OOD detection performance.

\paragraph{Optimal Trasnport.}
Optimal Transport (OT), a mathematical tool for measuring distributional discrepancies, has been widely adopted in various fields. In domain adaptation~\cite{DBLP:journals/pami/CourtyFTR17,DBLP:conf/aaai/Yang0023a,DBLP:conf/eccv/DamodaranKFTC18}, OT is used to align source and target domains. Wasserstein GAN~\cite{DBLP:conf/icml/ArjovskyCB17} leverages OT to measure distributional distance, using it as a loss function for training the generator. In object detection,~\cite{DBLP:conf/cvpr/GeLLYS21} applies OT to address the matching problem between anchors and ground truth. Several works explore OT in Out-of-Distribution (OOD) detection~\cite{DBLP:conf/ijcai/RenFXZ24,DBLP:conf/cvpr/Lu0ZZC23}. ~\cite{DBLP:conf/ijcai/RenFXZ24} introduces Partial OT to Open-Set Semi-Supervised Learning (OSSL), reallocating weights for unlabeled samples to align their distribution with the labeled data, enhancing interpretability in OOD detection. Similarly, \cite{DBLP:conf/cvpr/Lu0ZZC23} combines OT with clustering methods, leveraging distributional discrepancies between ID and OOD samples to assign unlabeled samples to appropriate clusters.

To the best of our knowledge, the application of OT in the zero-shot OOD detection realm remains unexplored. Existing methods use OT to model relationships between labeled training data (i.e.,~ID samples) and unlabeled data. In this work, we propose a zero-shot OOD detection framework that integrates OT with an OOD score, incorporating both semantic and distributional scores for improved detection performance.

\begin{table*}[t!]
    \centering
    \resizebox{1\linewidth}{!}{
    \begin{tabular}{lcccccccccc}
        \toprule
        \multirow{3}{*}{Prompt Templates}  & \multicolumn{10}{c}{\textbf{OOD Dataset}} \\
         ~& \multicolumn{2}{c}{iNaturalist} & \multicolumn{2}{c}{SUN} &\multicolumn{2}{c}{Places } & \multicolumn{2}{c}{Texture} & \multicolumn{2}{c}{\textbf{Average}}\\
        \cmidrule(lr){2-3} \cmidrule(lr){4-5} \cmidrule(lr){6-7} \cmidrule(lr){8-9} \cmidrule(lr){10-11}
        ~& \textbf{FPR95}↓ & \textbf{AUROC}↑ & \textbf{FPR95}↓ &  \textbf{AUROC}↑ & \textbf{FPR95}↓ &  \textbf{AUROC}↑ & \textbf{FPR95}↓ &  \textbf{AUROC}↑ & \textbf{FPR95}↓ &  \textbf{AUROC}↑ \\
        \midrule
         A bad photo of the [\texttt{CLASS}]. &11.87 & 97.48 & 19.63 & 95.47 &32.24 & 92.02 &42.62 & 91.20 &26.59 & 94.05 \\
         A tattoo of a [\texttt{CLASS}]. &6.35 & 98.46 & 14.63 & 96.72 &26.88 & 93.43 &47.62 & 88.48 &23.87 & 94.27 \\
         A bright photo of a [\texttt{CLASS}]. &8.75 & 97.89 & 16.29 & 96.17 &29.01 & 92.84 &43.23 & 90.25 &24.32 & 94.29 \\
         A photo of a clean [\texttt{CLASS}]. &8.22 & 98.04 & 17.53 & 95.91 &28.34 & 92.97 &42.39 & 90.19 &24.12 & 94.28 \\
         A photo of a dirty [\texttt{CLASS}]. &9.38 & 97.82 & 16.86 & 96.00 &31.06 & 92.16 &44.80 & 89.23 &25.53 & 93.80 \\
         A low resolution photo of a [\texttt{CLASS}]. &9.53 & 97.82 & 20.80 & 95.12 &30.77 & 92.41 &42.64 & 90.12 &25.94 & 93.87 \\
         A photo of a cool [\texttt{CLASS}]. &7.85 & 98.18 & 15.26 & 96.42 &28.61 & 92.72 &43.99 & 90.18 &23.93 & 94.38 \\
         A photo of a small [\texttt{CLASS}]. &8.49 & 97.86 & 17.41 & 95.91 &27.05 & 93.38 &41.95 & 90.80 &23.73 & 94.49 \\
         A dark photo of a [\texttt{CLASS}]. &8.63 & 98.02 & 17.86 & 96.05 &30.97 & 92.12 &44.41 & 89.41 &25.47 & 93.90 \\
         A blurry photo of a [\texttt{CLASS}]. &10.70 & 97.63 & 20.07 & 95.03 &31.53 & 91.19 &42.16 & 90.30 &26.12 & 93.54 \\
         A low resolution photo of a [\texttt{CLASS}]. &9.53 & 97.82 & 20.80 & 95.12 &30.77 & 92.41 &42.64 & 90.12 &25.94 & 93.87 \\
         A cropped photo of the [\texttt{CLASS}]. &9.99 & 97.70 & 20.36 & 95.37 &29.86 & 92.63 &43.16 & 90.19 &25.84 & 93.97 \\
         A good photo of a [\texttt{CLASS}]. &7.87 & 98.13 & 14.53 & 96.51 &29.06 & 92.59 &43.01 & 90.66 &23.62 & 94.47 \\
        
        [\texttt{CLASS}]. & 8.91 & 98.03 & 15.83 & 96.38 &24.26 & 93.82 &42.29 & 90.96 &22.82 & 94.80 \\
        A photo of a [\texttt{CLASS}].   & 7.91 & 98.16 &  16.18 & 96.15 &  27.43 & 93.30 &  43.07 & 90.34 &  23.65 & 94.49 \\ 
        \midrule
        The nice [\texttt{CLASS}].& 7.85 & 98.19 &  16.72 & 96.22 &  25.63 & 93.44 &  35.73 & 92.70 &  21.48 & 95.14 \\
        The nice cropped [\texttt{CLASS}]. & 11.41 & 97.43 &  20.41 & 95.57 &  29.40 & 92.76 &  41.49 & 91.61 &  25.68 & 94.34 \\
        The nice complete [\texttt{CLASS}]. &  6.63 & 98.44 &  16.50 & 96.38 &  23.18 & 94.13 &  39.18 & 91.86 &  21.37 & 95.20 \\
        \bottomrule
    \end{tabular}
    }
    \caption{Performance comparison with different prompts on the ImageNet-1K benchmark.}
    \label{ap:tab:prompt:1k}
\end{table*}

\begin{table*}[t!]
    \centering
    \resizebox{0.85\linewidth}{!}{
    \begin{tabular}{lccccccccccc}
        \toprule
        \multirow{3}{*}{ID Dataset} & \multirow{3}{*}{Method} & \multicolumn{10}{c}{OOD Dataset} \\
         & &  \multicolumn{2}{c}{iNaturalist} &  \multicolumn{2}{c}{SUN} &\multicolumn{2}{c}{Places } & \multicolumn{2}{c}{Texture} & \multicolumn{2}{c}{\textbf{Average}}\\
         \cmidrule(lr){3-4} \cmidrule(lr){5-6} \cmidrule(lr){7-8} \cmidrule(lr){9-10} \cmidrule(lr){11-12}
         & &  \textbf{FPR95}↓ & \textbf{AUROC}↑ & \textbf{FPR95}↓ &  \textbf{AUROC}↑ & \textbf{FPR95}↓ &  \textbf{AUROC}↑ & \textbf{FPR95}↓ &  \textbf{AUROC}↑ & \textbf{FPR95}↓ &  \textbf{AUROC}↑ \\

        \midrule
        \multirow{2}{*}{CUB-200} & MCM & 9.83 & 98.24 & 4.93 & 99.10 & 6.65 & 98.57 & 6.97 & 98.75 & 7.09 & 98.66 \\
        ~ & \textsc{OT-detector} & \textbf{6.73} & \textbf{98.64} & \textbf{0.02} & \textbf{99.99} & \textbf{0.52} & \textbf{99.85} & \textbf{0.02} & \textbf{99.99} & \textbf{0.17} & \textbf{99.95} \\
        \midrule
        \multirow{2}{*}{Stanford-Cars} & MCM & 0.05 & 99.77 & 0.02 & 99.95 & 0.24 & 99.89 & 0.02 & 99.96 & 0.08 & 99.89\\
        ~ &  \textsc{OT-detector} & \textbf{0.00} & \textbf{100.00} & \textbf{0.00} & \textbf{100.00} & \textbf{0.09} & \textbf{99.97} & \textbf{0.00} & \textbf{100.00} & \textbf{0.02} & \textbf{99.99} \\
        \midrule
        \multirow{2}{*}{Food-101} & MCM & 0.64 & 99.78 & 0.90 & 99.75 & 1.86 & 99.58 & \textbf{4.04} & \textbf{98.62} & 1.86 & 99.43 \\
        ~ & \textsc{OT-detector} & \textbf{0.22} & \textbf{99.96} &  \textbf{0.13} & \textbf{99.97} & \textbf{0.47} & \textbf{99.92} & 5.05 & 98.32 & \textbf{1.47} & \textbf{99.54} \\
       \midrule
        \multirow{2}{*}{Oxford-Pet} & MCM & 2.85 & 99.38 & 1.06 & 99.73 & 2.11 & 99.56 & 0.80 & 99.81 & 1.70 & 99.62\\
        ~ & \textsc{OT-detector} & \textbf{0.00} & \textbf{99.99} & \textbf{0.03} & \textbf{99.99} & \textbf{0.19} & \textbf{99.95} & \textbf{0.20} & \textbf{99.95} & \textbf{0.10} & \textbf{99.69} \\
        \midrule
        \multirow{2}{*}{ImageNet-10} & MCM & 0.12 & 99.80 & 0.29 & \textbf{99.79 }& 0.88 & \textbf{99.62} & \textbf{0.04} & \textbf{99.90} & 0.33 & \textbf{99.78}\\
        ~ & \textsc{OT-detector} & \textbf{0.08} & \textbf{99.82} & \textbf{0.22} & 99.74 & \textbf{0.72} & 99.56 & 0.07 & 99.84 & \textbf{0.27} & 99.74 \\
        \midrule
        \multirow{2}{*}{ImageNet-20} & MCM & 1.02 & 99.66 & \textbf{2.55} & \textbf{99.50} & 4.40 & 99.11 & 2.43 & \textbf{99.03} & 2.60 & 99.32\\

        ~ & \textsc{OT-detector} & \textbf{0.73} & \textbf{99.80} & 2.69 & 99.45 & \textbf{3.66} & \textbf{99.25} & \textbf{2.25} & 98.98 & \textbf{2.34} & \textbf{99.37} \\
        \midrule
        \multirow{2}{*}{ImageNet-100} & MCM & 18.13 & 96.77 & 36.45 & 94.54 & 34.52 & 94.36 & 41.22 & 92.25 & 32.58 & 94.48\\
        ~ & \textsc{OT-detector} & \textbf{4.87} & \textbf{98.99} & \textbf{8.83} & \textbf{97.86} & \textbf{17.27} & \textbf{96.53} & \textbf{24.40} & \textbf{95.78} & \textbf{13.84} & \textbf{97.29} \\
        \bottomrule
    \end{tabular}
    }
    \caption{Zero-shot OOD detection performance comparison on various ID datasets.}
    \label{cub}
\end{table*}

\subsection{Effect of Prompt Engineering}\label{ap:prompt}
We evaluate the impact of various prompt templates on the ImageNet-1K benchmark, as summarized in Table~\ref{ap:tab:prompt:1k}. Our method demonstrates consistent performance across a wide range of prompt templates, highlighting its robustness to prompt variations. Notably, the template ``the nice [\texttt{CLASS}].'' achieves the best performance.

To further investigate, we modify the template by adding ``cropped'' and ``complete'', resulting in ``the nice cropped [\texttt{CLASS}].'' and ``the nice complete [\texttt{CLASS}].'' templates. As shown in the last three rows of Table~\ref{ap:tab:prompt:1k}, using ``the nice cropped [\texttt{CLASS}].'' leads to a significant performance drop, whereas ``the nice complete [\texttt{CLASS}].'' improves performance further.
We attribute these results to the Semantic-aware Content Refinement (SaCR) module. SaCR retains the full object in ID images, while OOD images often contain only partial objects. Using ``cropped'' in the template increases the similarity between textual features and OOD images, causing a noticeable degradation in performance. Conversely, ``complete'' emphasizes the full object, enhancing the separation between ID and OOD samples and improving detection performance.

\subsection{Evaluation on Various ID Datasets} \label{ap:various_ID}
To further validate the efficacy of our method, we perform experiments on additional datasets. Following the setup in MCM~\cite{DBLP:conf/nips/MingCGSL022}, we use subsets of ImageNet, CUB-200 ~\cite{WahCUB_200_2011}, Stanford Cars ~\cite{DBLP:conf/iccvw/Krause0DF13}, Food-101 ~\cite{DBLP:conf/eccv/BossardGG14}, and Oxford Pets ~\cite{DBLP:conf/cvpr/ParkhiVZJ12} as ID data, while utilizing iNaturalist, SUN, Places, and Textures as OOD data. The results, summarized in Table~\ref{cub}, demonstrate that our method achieves an FPR95 of approximately $1\%$ and an AUROC near $100\%$ on CUB-200, Stanford Cars, Food-101, Oxford Pets, ImageNet-10, and ImageNet-20. On the ImageNet-100 dataset, our approach improves FPR95 by $57.52\%$ and AUROC by $2.89\%$, yielding substantial performance improvements. These findings underscore the robustness and generalizability of our method across various datasets.

\begin{table*}[t!]
    \centering
    \resizebox{0.8\linewidth}{!}{
    \begin{tabular}{lcccccccccc}
        \toprule
        \multirow{3}{*}{$\epsilon$}  & \multicolumn{10}{c}{\textbf{OOD Dataset}} \\
         ~& \multicolumn{2}{c}{iNaturalist} & \multicolumn{2}{c}{SUN} &\multicolumn{2}{c}{Places } & \multicolumn{2}{c}{Texture} & \multicolumn{2}{c}{\textbf{Average}}\\
        \cmidrule(lr){2-3} \cmidrule(lr){4-5} \cmidrule(lr){6-7} \cmidrule(lr){8-9} \cmidrule(lr){10-11}
        ~& \textbf{FPR95}↓ & \textbf{AUROC}↑ & \textbf{FPR95}↓ &  \textbf{AUROC}↑ & \textbf{FPR95}↓ &  \textbf{AUROC}↑ & \textbf{FPR95}↓ &  \textbf{AUROC}↑ & \textbf{FPR95}↓ &  \textbf{AUROC}↑ \\
        \midrule
        50 & 16.21 & 96.78 & 30.50 & 93.75 &40.03 & 90.86 &46.06 & 89.32 &33.20 & 92.68 \\
        60 & 11.55 & 97.49 & 24.52 & 94.79 &35.56 & 91.94 &44.88 & 89.78 &29.13 & 93.50 \\
        70 & 9.33 & 97.87 & 19.90 & 95.52 &31.02 & 92.68 &43.99 & 90.12 &26.06 & 94.05 \\
        80&  8.26 & 98.09 & 17.41 & 95.94 &29.23 & 93.10 &43.40 & 90.33 &24.58 & 94.36 \\
        90 & 7.91 & 98.16 & 16.18 & 96.15 &27.43 & 93.30 &43.07 & 90.34 &23.65 & 94.49 \\
        100 &  7.92 & 98.15 & 15.58 & 96.22 &27.05 & 93.36 &43.21 & 90.21 &23.44 & 94.49 \\
        110 & 8.08 & 98.14 & 15.52 & 96.22 &26.75 & 93.35 &43.78 & 89.97 &23.53 & 94.42 \\
        120 & 8.45 & 98.10 & 15.57 & 96.17 &26.74 & 93.29 &43.87 & 89.71 &23.66 & 94.32 \\
        130 &  8.97 & 97.98 & 15.83 & 96.11 &27.05 & 93.22 &44.33 & 89.41 &24.04 & 94.18 \\
        140 &  9.34 & 97.91 & 16.11 & 96.03 &27.39 & 93.14 &44.91 & 89.11 &24.44 & 94.05 \\
        150 &   9.75 & 97.83 & 16.34 & 95.94 &27.75 & 93.05 &45.55 & 88.81 &24.85 & 93.91 \\
        \bottomrule
    \end{tabular}
    }
    \caption{Detailed results on the impact of different $\epsilon$ with ImageNet-1k benchmark.}\vspace{-3mm}
    \label{ap:tab:epsi:1k}
\end{table*}

\subsection{More Experiments on Optimal Transport} \label{ap:ot}
This section investigates the impact of the hyperparameter $\epsilon$ in the entropy regularization term and the cost matrix $C$ in Eq.~\ref{eq14}.

\begin{figure}[t!]
    \centering
    \includegraphics[width=0.45\textwidth]{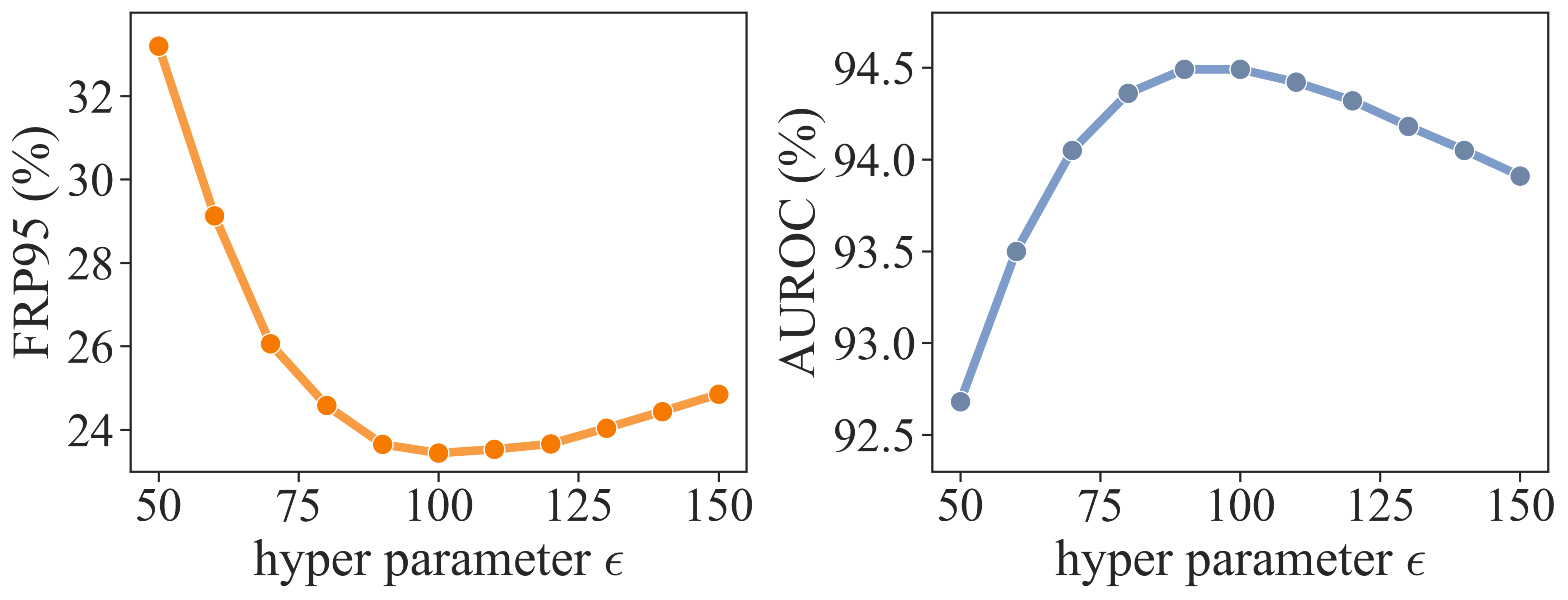}
    \caption{Analyses on the regularization parameter $\epsilon$,where
results are reported with ImageNet-1k benchmark.}
    \label{fig:ap:eps}
\end{figure}

\paragraph{Sensitivity Analysis on Hyperparameter $\epsilon$.} \label{ap:abeps}
The regularization parameter $\epsilon$ plays a crucial role in balancing the trade-off between smoothness and computational complexity, influencing the "softness" of the optimal transport solution~\cite{DBLP:conf/nips/Cuturi13}. As shown in Figure~\ref{fig:ap:eps}, the optimal value of $\epsilon$ for FPR95 is around $100$, where the minimum value is achieved. For AUROC, the maximum is attained at approximately $90$. Based on these observations, we select $\epsilon = 90$ for all subsequent experiments. Table~\ref{ap:tab:epsi:1k} further presents the effect of varying $\epsilon$ on four OOD datasets from the ImageNet-1K benchmark. The results indicate that the optimal $\epsilon$ varies slightly across different datasets, with $\epsilon = 90$ yielding satisfactory performance on all four datasets.

\paragraph{Choice of Cost Matrix $C$.}
We perform a comparative evaluation of cosine distance and L2 distance as the cost matrix on the ImageNet-1K benchmark, as summarized in Table~\ref{abaltion:cost}. The results show a significant performance improvement with cosine distance, suggesting that it is a more suitable measure for the distance between CLIP's textual and visual features.

\begin{table}[h]
    \centering
    
    \resizebox{0.7\linewidth}{!}{
    \begin{tabular}{ccc}
        \toprule
        Metric & \textbf{FPR95}↓ &  \textbf{AUROC}↑  \\
        \midrule
        L2 distance &   25.57 & 94.15 \\
        Cosine distance & \textbf{23.65} &\textbf{94.49}\\

        \bottomrule
    \end{tabular}
    }
    \caption{Different cost metrics evaluated with the ImageNet-1K benchmark.}
    \label{abaltion:cost}
\end{table}

\subsection{Additional Experiments on the SaCR Module} \label{ap:sacr}
This section discusses the number of views $N$ in \textbf{view augmentation}, the selection of the confidence function, and the constraint on maintaining consistency in view labels during \textbf{view selection}.

\paragraph{Sensitive Analysis of the View Number $N$.} \label{ap:N}
We conduct experiments on the ImageNet-1K benchmark to evaluate the impact of the number of views $N$ on zero-shot OOD detection performance, as shown in Table~\ref{ap:tab:abn}. Here, $N=1$ denotes the use of only the original image. The augmentation step primarily affects the degree of decoupling of the original image content. Increasing $N$ generally enhances the ability to capture more diverse object parts that convey semantic information from the original image, thus influencing the quality of the refined features obtained through fusion. Table~\ref{ap:tab:abn} demonstrates that using only the original image ($N=1$) yields the lowest performance. As $N$ increases, performance steadily improves, indicating that additional views facilitate the extraction of more local features, which helps to capture more comprehensive and semantically relevant regions in OOD samples that correspond to the ID samples.


\begin{table}[t]
    \centering
    
    \resizebox{0.9\linewidth}{!}{
    \begin{tabular}{cccc}
        \toprule
        SaCR & Label Consistency & \textbf{FPR95}↓ &  \textbf{AUROC}↑  \\
        \midrule
        \ding{55} &  \ding{55} & 32.65 & 94.96 \\
        \ding{51} &  \ding{55} & 24.42 & 95.82 \\
        \ding{51} &   \ding{51} &\textbf{22.60} &\textbf{96.15}\\

        \bottomrule
    \end{tabular}
    }
    \caption{Effect of the Semantic Guidance with ImageNet-100/ImageNet-10 as ID/OOD data.}
    \label{abaltion:hardfc}
\end{table}

\begin{table*}[t]
    \centering
    \resizebox{0.9\linewidth}{!}{
    \begin{tabular}{ccccccccccc}
        \toprule
        \multirow{3}{*}{$N$ Views}  & \multicolumn{10}{c}{\textbf{OOD Dataset}} \\
         ~& \multicolumn{2}{c}{iNaturalist} & \multicolumn{2}{c}{SUN} &\multicolumn{2}{c}{Places } & \multicolumn{2}{c}{Texture} & \multicolumn{2}{c}{\textbf{Average}}\\
        \cmidrule(lr){2-3} \cmidrule(lr){4-5} \cmidrule(lr){6-7} \cmidrule(lr){8-9} \cmidrule(lr){10-11}
        ~& \textbf{FPR95}↓ & \textbf{AUROC}↑ & \textbf{FPR95}↓ &  \textbf{AUROC}↑ & \textbf{FPR95}↓ &  \textbf{AUROC}↑ & \textbf{FPR95}↓ &  \textbf{AUROC}↑ & \textbf{FPR95}↓ &  \textbf{AUROC}↑ \\
        \midrule
        1 & 10.44 & 97.69 & 22.72 & 95.07 &35.16 & 91.66 &49.84 & 87.25 &29.54 & 92.92 \\
        8 & 10.65& 97.61 & 23.31 & 94.94 & 34.05 & 92.07 & 46.63 & 89.16 & 28.66 & 93.45 \\
        32 & 9.21 & 97.91 & 20.08 & 95.50 & 31.39 & 92.70 & 44.11 &  89.97 & 26.20& 94.02 \\
        64 & 8.18 & 98.11 & 18.04 & 95.91 & 29.80 & 93.14 & 44.72 & 89.97 & 25.18 & 94.28 \\
        128 & 8.20 & 98.11& 17.61 & 95.92 & 29.23 & 93.13 & 44.31 & 90.15 & 24.84 & 94.33 \\
        256 & \textbf{7.91} & \textbf{98.16} &\textbf{16.18} & \textbf{96.15} & \textbf{27.43} & \textbf{93.30} & \textbf{43.07} & \textbf{90.34} & \textbf{23.65} & \textbf{94.49} \\    
        
        \bottomrule
    \end{tabular}
    }
    \caption{Effect of the views number $N$, where experiments are conducted with ImageNet-1K benchmark.}
    \label{ap:tab:abn}
\end{table*}

\paragraph{Choice of Confidence Function.} 
We investigate the impact of various confidence functions on view selection, as shown in Table~\ref{ap:tab:cf}. The primary confidence functions considered include \textit{min-entropy}, \textit{min-margin}, and \textit{max-margin}. Specifically, the min-entropy method computes the self-entropy for each view and selects the view with the lowest entropy value, yielding satisfactory performance. Among the evaluated strategies, the max-margin approach demonstrates the best performance.

\label{ap:cf}
\begin{table}[t]
    \centering
    \resizebox{0.7\linewidth}{!}{
    \begin{tabular}{ccc}
        \toprule
        \textbf{Confidence Function}  & \textbf{FPR95}↓ &  \textbf{AUROC}↑ \\
        \midrule
        Min-Entropy & 24.16& 94.35 \\
        Min-Margin & 28.51 & 93.39 \\
        Max-Margin  & \textbf{23.65} & \textbf{94.49} \\
        \bottomrule
    \end{tabular}
    }
    \caption{Performance comparison of different confidence functions, evaluated with the ImageNet-1K benchmark.}
    \label{ap:tab:cf}
\end{table}

\begin{figure}[h]
    \centering
    \begin{subfigure}[b]{0.235\textwidth}
        \centering
        \includegraphics[width=\textwidth]{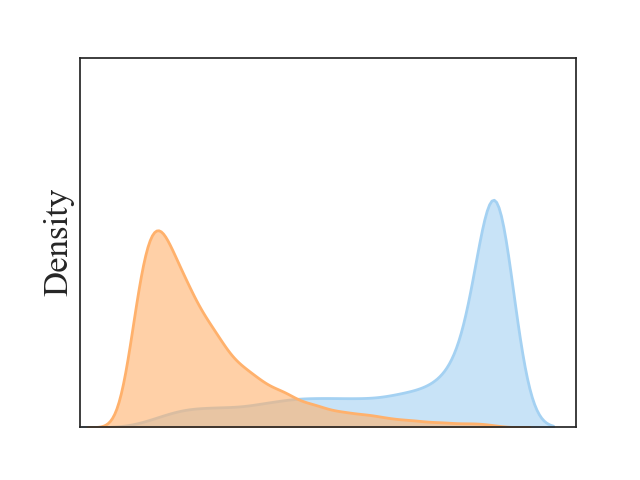}\vspace{-2mm}
        \caption{w/o SaCR}
    \end{subfigure}
    \begin{subfigure}[b]{0.235\textwidth}
        \centering
        \includegraphics[width=\textwidth]{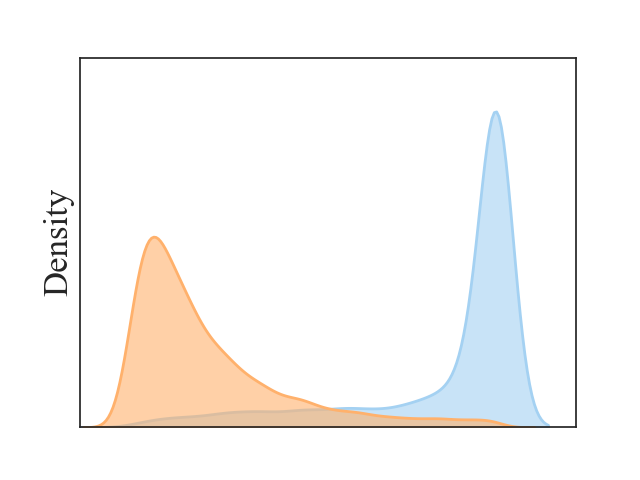}\vspace{-2mm}
        \caption{with SaCR}
    \end{subfigure}\vspace{-2mm}
    \caption{Score distribution of the ID (blue) and OOD (orange) score w/o SaCR (left) and with SaCR (right).}
    \label{fig:ap:density}
\end{figure}

\paragraph{Impact of Maintaining View Label Consistency.}
We conduct ablation experiments using ImageNet-100/ImageNet-10 as the ID/OOD data to validate the effectiveness of maintaining label consistency in view selection within SaCR. The experimental results, presented in Table~\ref{abaltion:hardfc}, show that view selection based on confidence scores leads to significant performance improvements even without label consistency. Incorporating label consistency into SaCR further enhances the results, achieving a notable reduction in FPR95 by $30.78\%$. This demonstrates the effectiveness of SaCR in addressing the challenging problem of hard OOD detection. To further illustrate the impact of SaCR, we compare the density plots of ID and OOD scores before and after applying SaCR, as shown in Figure~\ref{fig:ap:density}. The results indicate that SaCR reduces sensitivity to threshold selection when distinguishing between ID and OOD samples. Additionally, ID scores are significantly elevated, highlighting a clearer separation between ID and OOD distributions.

\subsection{Evaluation with Different Versions of CLIP
} \label{ap:backbone}
In this section, we evaluate the performance of \textsc{OT-detector} using different CLIP versions and compare it with two competitive methods, EOE and NegLabel. Table~\ref{ap:tab:backbone} presents the results on the ImageNet-1K benchmark for various CLIP versions, including ViT-B/32, ViT-B/16, and ResNet50. Our method achieves the best performance across both convolutional and attention-based vision encoders, demonstrating the strong generalization capability of our approach across different visual architectures.

\begin{table*}[t!]
    \centering
    \resizebox{0.8\linewidth}{!}{
    \begin{tabular}{lcccccccccc}
        \toprule
        \multirow{3}{*}{Method (CLIP version)}  & \multicolumn{10}{c}{\textbf{OOD Dataset}} \\
         ~& \multicolumn{2}{c}{iNaturalist} & \multicolumn{2}{c}{SUN} &\multicolumn{2}{c}{Places } & \multicolumn{2}{c}{Texture} & \multicolumn{2}{c}{\textbf{Average}}\\
        \cmidrule(lr){2-3} \cmidrule(lr){4-5} \cmidrule(lr){6-7} \cmidrule(lr){8-9} \cmidrule(lr){10-11}
        ~& \textbf{FPR95}↓ & \textbf{AUROC}↑ & \textbf{FPR95}↓ &  \textbf{AUROC}↑ & \textbf{FPR95}↓ &  \textbf{AUROC}↑ & \textbf{FPR95}↓ &  \textbf{AUROC}↑ & \textbf{FPR95}↓ &  \textbf{AUROC}↑ \\
        \midrule

        EOE (ResNet50) & 14.44 & 97.05 & 24.85 & 94.82 & 38.89 & 90.64 & 59.73 & 85.44 & 34.47 & 91.98 \\
        NegLabel (ResNet50)  & \textbf{2.88} & \textbf{99.24} & 26.51 & 94.54 & 42.60 & 89.72 & 50.80 & 88.40 & 30.70 & 92.97 \\
        \textsc{OT-detector} (ResNet50) &12.45 & 97.19 & \textbf{21.21} & \textbf{95.09} & \textbf{34.87} & \textbf{91.28} & \textbf{49.86} & \textbf{88.43} & \textbf{29.60} & \textbf{93.00} \\

        \midrule

        EOE (VIT-B/32) & 14.97 & 97.03 & 21.88 & 95.40 & 29.91 & 92.97 & 60.10 & 84.61 & 31.72 & 92.50 \\
        NegLabel (ViT-B/32) & \textbf{3.73} & \textbf{99.11} & 22.48 & 95.27 & 34.94 & 91.72 & 50.51 & 88.57 & 27.92 & 93.67 \\
        \textsc{OT-detector} (Vit-B/32) & 10.16 & 97.76 & \textbf{19.26} & \textbf{95.53} & \textbf{27.80} & \textbf{93.20} & \textbf{44.89} & \textbf{89.59} & \textbf{25.53} & \textbf{94.02}  \\

        \midrule

        EOE (Vit-B/16) & 12.29 & 97.52 & 20.40 & 95.73 & 30.16 & 92.95 & 57.53 & 85.64 & 30.09 & 92.96 \\
        NegLabel (Vit-B/16) & \textbf{1.91} & \textbf{99.49} & 20.53 & 95.49 & 35.59 & 91.64 & 43.56 & 90.22 & 25.40 & 94.21 \\
        \textsc{OT-detector} (Vit-B/16) & 7.91 & 98.16 &\textbf{16.18} & \textbf{96.15} & \textbf{27.43} & \textbf{93.30} & \textbf{43.07} & \textbf{90.34} & \textbf{23.65} & \textbf{94.49}  \\
        \bottomrule
    \end{tabular}
    }
    \caption{Additional experimental results using different versions of CLIP on the ImageNet-1K benchmark.}
    \label{ap:tab:backbone}
\end{table*}

\begin{table*}[htbp]
    \centering
    \resizebox{0.8\linewidth}{!}{
    \begin{tabular}{lcccccccccc}
        \toprule
          & \multicolumn{10}{c}{\textbf{OOD Dataset}} \\
         & \multicolumn{2}{c}{iNaturalist} & \multicolumn{2}{c}{SUN} &\multicolumn{2}{c}{Places } & \multicolumn{2}{c}{Texture} & \multicolumn{2}{c}{\textbf{Average}}\\
        \cmidrule(lr){2-3} \cmidrule(lr){4-5} \cmidrule(lr){6-7} \cmidrule(lr){8-9} \cmidrule(lr){10-11}
        Batch& \textbf{FPR95}↓ & \textbf{AUROC}↑ & \textbf{FPR95}↓ &  \textbf{AUROC}↑ & \textbf{FPR95}↓ &  \textbf{AUROC}↑ & \textbf{FPR95}↓ &  \textbf{AUROC}↑ & \textbf{FPR95}↓ &  \textbf{AUROC}↑ \\
        \midrule
        16 & 51.97 & 89.85 & 62.12 & 86.72 &  63.48 & 84.81  & 63.85 & 84.87 &60.35 & 86.56 \\
        64 & 32.00 & 93.37& 47.08 &  90.44 & 52.08 & 88.18 & 53.69 & 86.49  &46.21 &89.62 \\
        128 & 26.58 & 94.51 &43.48 & 91.22 &49.38 & 88.87  &51.60 & 86.73 &42.76 & 90.33 \\
        256&  23.12 & 95.26 &38.82 & 92.20&46.21 & 89.67 &50.16 & 87.06  & 39.58 & 91.05 \\
        512 & 18.89 & 95.89  &35.19 & 93.07&42.04 & 90.65 &48.78 & 87.69 & 36.22 & 91.82 \\
        1024 & 15.46 & 96.61  &29.22 & 93.82 &38.43 & 91.24&47.59 & 88.15 & 32.67 & 92.45 \\
        2048 & 12.46 & 97.24 &22.82 & 94.82 &34.09 & 92.06&46.49 & 88.66 & 28.96 & 93.19 \\
        5096 & 9.51 & 97.80 &19.07 & 95.55  &30.85 & 92.71  &45.71 & 89.13 & 26.28 & 93.79 \\
        ALL &  \textbf{7.91} & \textbf{98.16} &\textbf{16.18} & \textbf{96.15} & \textbf{27.43} & \textbf{93.30} & \textbf{43.07} & \textbf{90.34} & \textbf{23.65} & \textbf{94.49} \\
        \bottomrule
    \end{tabular}
    }
    \caption{Effect of different batch sizes with ImageNet-1K benchmark.}
    \label{ap:tab:batch:1k}
\end{table*}

\subsection{Limitation} \label{ap:limit}
\paragraph{Sensitivity to Different Batch Sizes.}
Due to the inherent limitations of Optimal Transport (OT), \textsc{OT-detector} requires a sufficient number of test features to quantify the distributional discrepancy between the test features and the ID textual features using OT, and subsequently identify OOD samples based on the OOD score $S_\text{OT}$. It is important to note that our method does not rely on the labels of the test data. Instead, it only requires the cosine distance between batches of test data features and the ID label features. OT optimization is then performed based on these cosine distances. Therefore, unlike methods such as MCM and NegLabel, our approach cannot provide real-time evaluations for individual test samples.

The use of OT also introduces sensitivity to batch size, as larger batches allow the observed distribution to better approximate the true distribution of the test data. To evaluate this, we conduct experiments with varying batch sizes using the ImageNet-1K benchmark, as shown in Table~\ref{ap:tab:batch:1k}. The results show that as the batch size increases, the performance of our method improves, reaching its peak when the entire test dataset is utilized.

In future work, we aim to further explore the integration of OT with zero-shot OOD detection to improve performance with smaller batch sizes, thereby enhancing the practicality and efficiency of the approach in real-world applications.

\paragraph{Requirement for Sufficient ID Labels.}
As shown in Table~\ref{hardood}, \textsc{OT-detector} achieves an FPR95 of $3.06\%$ and an AUROC of $99.17\%$ when using ImageNet-10 as ID data and ImageNet-100 as OOD data. While this performance is slightly lower than that of MCM, it still represents a commendable result. We attribute this to the fact that when ImageNet-10 is used as ID data, the number of ID labels is significantly smaller than the number of categories in the OOD data. In such an imbalanced scenario, the semantic information provided by the ID labels is insufficient to generate sufficiently effective views of the OOD data with complex categories, thereby limiting the effectiveness of SaCR. However, when a sufficient number of ID categories is available, such as when using ImageNet-100/ImageNet-10 as ID/OOD data, our method outperforms others, even significantly surpassing NegLabel.


\subsection{Discussion on NegLabel} \label{ap:neglabel}

As shown in Table~\ref{Image1K}, when compared to NegLabel~\cite{DBLP:conf/iclr/Jiang000LZ024}, our approach exhibits slightly inferior performance on the iNaturalist dataset. However, it shows substantial improvements over NegLabel on the SUN and Places datasets, resulting in superior overall performance. It is important to note that for the ImageNet-1K benchmark, NegLabel utilizes ten times the number of ID labels as negative labels.

The superior performance of NegLabel on the iNaturalist dataset can be primarily attributed to the nature of iNaturalist, which consists mostly of images of natural plants. In contrast, ImageNet-1K covers a much broader range of categories, predominantly focusing on food and animals. Therefore, iNaturalist differs significantly from ImageNet-1K in terms of category composition. As a result, the negative labels introduced by NegLabel are particularly effective in describing the iNaturalist dataset, even enabling the identification of the ground-truth label for a given image. On the other hand, datasets such as SUN and Places exhibit greater semantic similarity to ImageNet-1K, leading to less effective performance for NegLabel. While our method incorporates additional distributional discrepancy, it is challenging to outperform NegLabel on the iNaturalist dataset. Nevertheless, for the other three OOD datasets, our method demonstrates better performance.

Additionally, as reported in NegLabel, its performance is highly sensitive to the prompt template. Arbitrary modifications to the prompt template can lead to a significant performance drop. Through prompt engineering, NegLabel identifies the template "the nice [\texttt{CLASS}]" as yielding the best results. In contrast, as discussed in Section~\ref{ap:prompt}, \textsc{OT-detector} consistently delivers stable performance across various templates, demonstrating its strong robustness to variations in the prompt template.
